\documentclass{article} 
\usepackage{iclr2026_conference,times}

\usepackage{amsmath,amsfonts,bm}

\def\eqref#1{equation~\ref{#1}}

\def\1{\bm{1}}

\DeclareMathAlphabet{\mathsfit}{\encodingdefault}{\sfdefault}{m}{sl}
\SetMathAlphabet{\mathsfit}{bold}{\encodingdefault}{\sfdefault}{bx}{n}

\usepackage{hyperref}
\usepackage{url}

\usepackage[utf8]{inputenc} 
\usepackage[T1]{fontenc}    
\usepackage{hyperref}       
\usepackage{url}            
\usepackage{booktabs}       
\usepackage{amsfonts}       
\usepackage{nicefrac}       
\usepackage{microtype}      
\usepackage{xcolor}         
\usepackage{graphicx}
\usepackage{wrapfig}
\usepackage{subcaption}
\usepackage{textcomp}
\usepackage{amsmath}
\usepackage{longtable}

\title{Unveiling the Role of Data Uncertainty \\ in Tabular Deep Learning}

\author{%
  Nikolay Kartashev \\
  HSE University, Yandex\\
  \And
  Ivan Rubachev \\
  Yandex, HSE University\\
  \And
  Artem Babenko \\
  Yandex, HSE University \\
}

\iclrfinalcopy

\begin{document}

\maketitle

\begin{abstract}
Recent advancements in tabular deep learning have demonstrated exceptional practical performance, yet the field often lacks a clear understanding of why these techniques actually succeed. To address this gap, our paper highlights the importance of the concept of data uncertainty for explaining the effectiveness of the recent tabular DL methods. 
In particular, we reveal that the success of many beneficial design choices in tabular DL, such as numerical feature embeddings, retrieval-augmented models and advanced ensembling strategies, can be largely attributed to their implicit mechanisms for managing high data uncertainty. By dissecting these mechanisms, we provide a unifying understanding of the recent performance improvements. 
Furthermore, the insights derived from this data-uncertainty perspective directly allowed us to develop more effective numerical feature embeddings as an immediate practical outcome of our analysis. Overall, our work paves the way to foundational understanding of the benefits introduced by modern tabular methods that results in the concrete advancements of existing techniques and outlines future research directions for tabular DL.
\end{abstract}

\section{Introduction}

Deep learning for tabular data is experiencing rapid progress, with new models and training approaches constantly improving performance across a wide range of tasks \citep{gorishniy2022embeddings, hollmann2022tabpfn, gorishniy2023tabr, ye2024modern, holzmüller2024better, gorishniy2024tabm, hollmann2025accurate}. However, a clear understanding of why the proposed techniques are effective mostly lags behind their empirical success. New methods are often introduced based primarily on empirical observations or by repurposing ideas from other domains, often without a thorough investigation of their effectiveness within the tabular data context. This gap between performance and understanding can hinder the future progress, potentially leading to a more trial-and-error research style.

\begin{figure*}[h!]
    \centering
    \includegraphics[width=\linewidth]{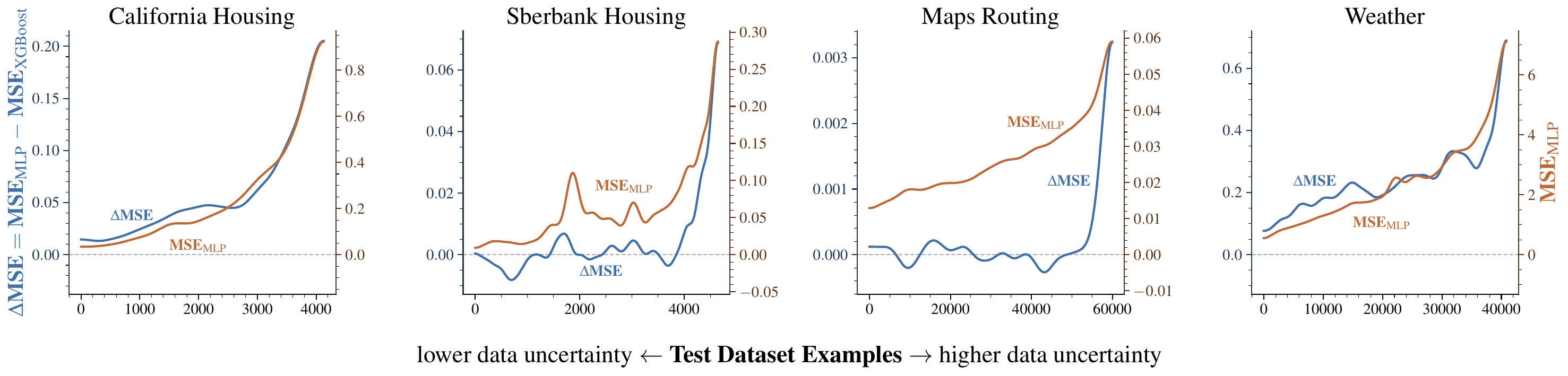}
    \caption{The performance differences measured by MSE between MLP and XGBoost are shown in blue, and absolute values of MSE for MLP are shown in brown. The figure has two separate vertical axes: the left ones correspond to $\Delta$MSE, and the right ones correspond to absolute MSE. Differences are reported for each test example, sorted from left to right by increasing data uncertainty. On four datasets, especially on Sberbank Housing and Maps Routing, MLP has significantly worse performance compared to XGBoost on high data uncertainty points. On Weather, this phenomenon is more subtle, but in the region of high uncertainty, the MSE difference still grows faster than MSE itself.}
    \label{fig:xgb-compare}
\end{figure*}

In this paper, we address this gap by leveraging the concept of \textit{data (aleatoric) uncertainty} \citep{gal2016uncertainty} as an informative tool for analyzing and understanding the effectiveness of recent tabular DL methods. Our rough motivation to employ data uncertainty originates from the intuition that, compared to computer vision or NLP, tabular problems often possess unobserved confounding variables and inherent noise in target labels that makes it challenging for models --- and even human experts --- to achieve perfect predictive accuracy due to irreducible noise or ambiguity. Furthermore, much indirect evidence also indicates high data uncertainty being a significant factor in tabular learning. For instance, the critical reliance of tabular DL performance on early stopping hints at the presence of significant noise in the labels \citep{baek2024sam}. While any single point above does not strictly necessitate that data uncertainty is an important constituent of a typical tabular problem, accumulated evidence from them convinced us to take a closer look at this tool. Interestingly, in the preliminary experiments, we compared a simple MLP model to the leading GBDT implementation in terms of performance on datapoints with different data uncertainty. For several datasets, \autoref{fig:xgb-compare} shows that GBDTs perform much better specifically in the high data uncertainty niche, which indicates that the ``DL vs GBDT'' battle probably unfolds mostly there.

In light of this, we systematically investigate several influential findings from the tabular DL literature --- specifically, numerical feature embeddings \citep{gorishniy2022embeddings}, retrieval-augmented models \citep{ye2024modern}, and advanced ensembling strategies \citep{gorishniy2024tabm} --- in terms of their ability to handle data uncertainty. Our investigation reveals a compelling and consistent pattern: a significant portion of these techniques' success can be attributed to their implicit yet effective mechanisms for managing datapoints characterized by high data uncertainty. In more practical terms, the techniques often appear to be disproportionately more beneficial on the highly uncertain datapoints. To obtain a clearer understanding, for each technique we dissect the specific mechanisms that enable the management of these challenging samples.
As a direct practical outcome of our analysis, we employ the proposed data-uncertainty viewpoint to develop a novel numerical feature embedding scheme that is more effective than existing alternatives.

Overall, the main contributions of our paper are:

\begin{itemize}

\item We introduce data uncertainty as an important tool to analyze the performance of various tabular learning methods and demonstrate that it provides an informative viewpoint on the methods' advantages.

\item We explain the specific mechanism behind the effectiveness of several recent tabular DL techniques, including numerical feature embeddings, retrieval-augmented models, and parameter-efficient ensembling.

\item We leverage the insights from our uncertainty-driven analysis to design a new, more effective numerical feature embedding scheme.

\end{itemize}

\section{Related work}

\textbf{Tabular Learning.} The field of tabular data modeling has seen a significant transformation in recent years, with deep learning approaches increasingly challenging the long-held supremacy of traditional ``shallow'' decision tree-based ensembles, such as Gradient Boosting Decision Trees (GBDTs). In particular, the most recent DL models \citep{gorishniy2023tabr, holzmüller2024better, ye2024modern, gorishniy2024tabm, hollmann2025accurate} have compellingly demonstrated performance on par with, or even higher, than that of leading GBDT implementations like CatBoost \citep{prokhorenkova2018catboost}, LightGBM \citep{ke2017lightgbm}, and XGBoost \citep{chen2016xgboost}. This practical success of tabular DL is a consequence of recent research efforts on novel architectures \citep{gorishniy2021revisiting, somepalli2021saint, holzmüller2024better, gorishniy2022embeddings, ye2024modern, gorishniy2023tabr, gorishniy2024tabm}, specialized regularizations and learning protocols \citep{bahri2021scarf, rubachev2022revisiting, jeffares2023tangos, thimonier2024t}. However, despite this plethora of empirically successful techniques, an understanding of the core principles behind their effectiveness often remains unclear, highlighting a need for deeper analysis of their underlying mechanisms and comparative strengths.

\textbf{Analysis in Tabular DL.} Several recent papers also contribute to a deeper understanding of the strengths and limitations of tabular DL methods. \citet{grinsztajn2022why} identifies the reasons why tabular DL can be inferior to GBDT methods on certain datasets and formulates the desiderata for tabular DL methods, in particular, the ability to model irregular target dependencies. \citet{mcelfresh2023neural} analyses the properties of tabular datasets that can be indicative of whether DL or GBDT models should be preferred. \citet{rubachev2024tabred} re-evaluates a large number of recent design choices in tabular DL on a more realistic benchmark and reveals that many of them are not robust to a temporal train--test shift. In contrast to these dataset-centric or technique-centric investigations, our paper proposes a more fine-grained sample-wise analysis, examining performance at the individual datapoint resolution.

\textbf{Uncertainty Estimation.} Uncertainty is a core concept in machine learning, reflecting potential inaccuracies in model predictions \citep{gal2016uncertainty}. Our work concentrates on data (aleatoric) uncertainty, which is inherent to the data itself --- typically arising from noise or randomness in the underlying process being modelled (e.g., noisy features or targets). Importantly, this type of uncertainty is independent of the specific model used and represents a fundamental limit on predictability for a given data point. However, while the underlying noise is fixed, models do differ in how effectively they cope with it. As we show, different tabular DL models exhibit varying levels of robustness to this inherent noise, meaning some are better than others at yielding robust predictions despite the intrinsic ambiguity present in certain samples.

\section{Preliminaries}
\label{preliminaries}

In this section, we outline the core concepts and notation required to analyze how different tabular deep learning approaches interact with data uncertainty.

The concept of data uncertainty arises when the target dependency is not deterministic, and instead the relationship between features $x$ and target variable $y$ is set by a non-degenerate conditional distribution $p(y \mid x)$. This can occur for several reasons, for example, in some cases, the label-collection procedure can be noisy due to mistakes of human annotators or due to imperfect sensors. In other cases, features in the dataset do not strictly determine the target variable; for example, it is impossible to know an exact temperature in a given city by its longitude and latitude alone. In these cases, the labels provided in a given dataset are samples from this conditional distribution $p(y \mid x)$.

The distribution $p(y \mid x)$ may have different characteristics for different $x$. For example, measurements of a target quantity could be exact for one set of $x$ and only an approximation for another. The concept of \textbf{data uncertainty} is used to measure how ``spread out'' the distributions $p(y \mid x)$ are for different samples of the data.

Informally, \textbf{data uncertainty} quantifies the inherent noise, particularly label noise, present in the data. While the loss in performance that comes from label noise in the test data is irreducible, it has been shown \citep{tanaka2018jointoptimizationframeworklearning} that when training on noisy labels, the performance of a model decreases even when measured on a clean test set. Different approaches and models can be influenced by this decrease to a different degree, which gives rise to a phenomenon we study in this paper. 

Our analysis focuses on regression tasks, in which data uncertainty for sample $i$ is defined as the variance of the distribution $p(y_i \mid x_i)$. In \autoref{sec:embeddings}, \autoref{sec:mnca}, and \autoref{sec:tabm}, we show how increases in performance for recent methods in tabular deep learning are disproportionately larger on samples where data uncertainty is higher. 

\subsection{Data Uncertainty Estimation}

A large part of our analysis relies on the empirical estimates of data uncertainty. When estimating data uncertainty, we assume the following probabilistic model: 

\begin{equation}
y_i = f(x_i) + e^{g(x_i)} \cdot \mathcal{N}(0, 1) \label{eq:unc_model}
\end{equation}

Here $\left(x_i, y_i\right)$ represents a datapoint, where $x_i$ is the feature set and $y_i$ is the target variable. The deterministic function $f(x_i)$ is the noiseless part of the target dependency, while $e^{2 \cdot g(x_i)}$ represents the data uncertainty\footnote{This parameterization ensures that data uncertainty values are always positive.}. To obtain our estimates of data uncertainty, we train a machine learning model to predict both $\left(f(x_i), g(x_i)\right)$ for any given datapoint $x_i$. The model is trained by maximizing the likelihood of the observed target values under a distribution $\mathcal{N}(f(x_i), e^{2 \cdot g(x_i)})$. This estimation approach aligns with the recent works \citep{duan2020ngboost, malinin2020uncertainty}. In our experiments, we use CatBoost models \citep{prokhorenkova2018catboost} to estimate the data uncertainty values, though any suitable deep learning model could be employed as well.

\begin{figure*}[h!]
    \centering
    \includegraphics[width=\linewidth]{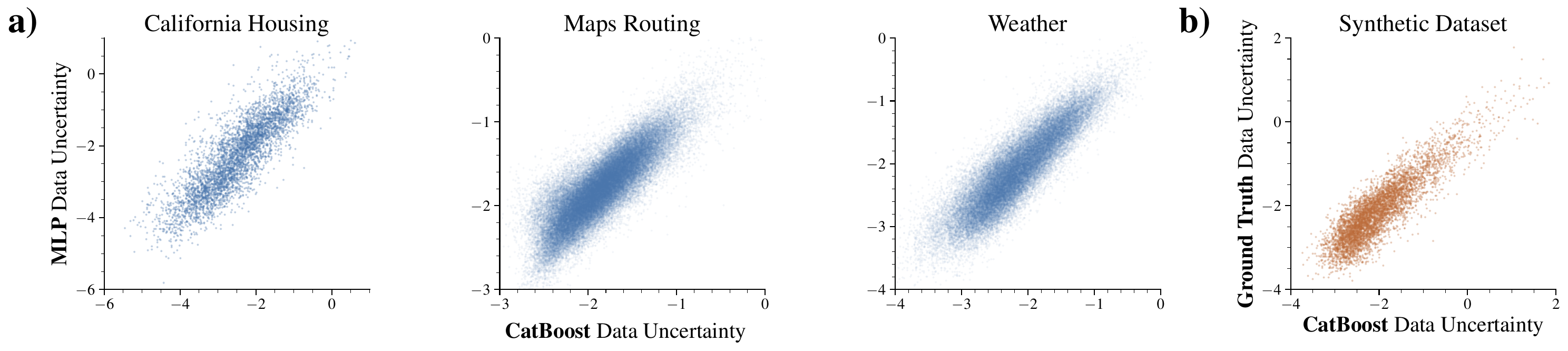}
    \caption{
    \textbf{a)} The scatterplots show a high correlation between the log-transformed data uncertainty estimates from CatBoost and an MLP. \textbf{b)} The scatterplot shows the relationship between CatBoost's log-estimated data uncertainty and the logarithms of the true data uncertainty values used for target sampling in the controlled synthetic experiment.}
    \label{fig:unc-compare}
    \phantomsubcaption\label{fig:unc-compare:a}
    \phantomsubcaption\label{fig:unc-compare:b}
    \vspace{-10pt}
\end{figure*}

Crucially, we find that the predicted data uncertainty values $e^{2\cdot g(x_i)}$ are relatively stable regardless of the specific estimator model used. For instance, see \autoref{fig:unc-compare:a} which demonstrates the strong correlations of uncertainty estimates produced by CatBoost and MLP on the California Housing, Maps Routing and Weather datasets. Moreover, the effects we analyze in \autoref{sec:embeddings}, \autoref{sec:mnca} and \autoref{sec:tabm} stay the same regardless of which uncertainty estimation method we use, as shown in \autoref{sec-A:consistency}. 

To further confirm the reliability of our empirical data uncertainty estimates, we generate a synthetic dataset where both functions $f(\cdot)$ and $g(\cdot)$ from \autoref{eq:unc_model} are randomly initialized MLPs. \autoref{fig:unc-compare:b} shows that the uncertainty values predicted by CatBoost closely match the ground-truth uncertainty values on this synthetic data. Moreover, we compare different tabular DL methods on datasets described in \autoref{sec:saw_spec} and \autoref{sec:mlp_synth} and demonstrate that an increase in performance coincides with high data uncertainty even when the true values of data uncertainty are known and no estimation is used in the analysis. 

To analyze how various tabular DL methods perform on datapoints with different levels of data uncertainty, we draw \textit{uncertainty plots} as follows. First, we obtain an estimate of data uncertainty for each test sample in the dataset as described above. Then, test datapoints are sorted by increasing values of their estimated data uncertainty (or true data uncertainty for the synthetic dataset described in \autoref{sec:mlp_synth}). For each datapoint, we then compute the 
performance (in terms of Mean Squared Error) of the simple MLP model and the method whose behavior we aim to analyze. Then, for visualization purposes, we additionally smooth the resulting curve (see the details in \autoref{sec-A:uncertainty-details}). As an illustrative example, \autoref{fig:xgb-compare} shows the uncertainty plot that compares MLP and XGBoost on several datasets and reveals that MLP has much worse performance on datapoints of high data uncertainty.

\section{Synthetic Data}
\label{sec:saw}

For real datasets, it is not entirely clear whether our estimation of data uncertainty reliably represents the true data uncertainty. For this reason, we first use synthetic datasets to demonstrate how the performance of several tabular DL methods changes at different levels of data uncertainty.

\subsection{Synthetic dataset with data uncertainty parameterized by an MLP}
\label{sec:mlp_synth}

The first dataset we use is obtained in the following way. First, we perform i.i.d. sampling from the $20$-dimensional standard Gaussian distribution to obtain the feature vectors $x_i$. Then, we produce the target variables as $f(x_i) + e^{g(x_i)} \cdot \mathcal{N}(0, 1)$, where both $f(\cdot)$ and $g(\cdot)$ are parameterized as randomly initialized MLPs. As a result of this procedure, we obtain a dataset for which data uncertainty exists and we know its true values for each sample. For full details regarding the creation of this dataset, see \autoref{sec-A:synthetic-old}. 

\begin{figure*}[h!]
    \centering
    \includegraphics[width=\linewidth]{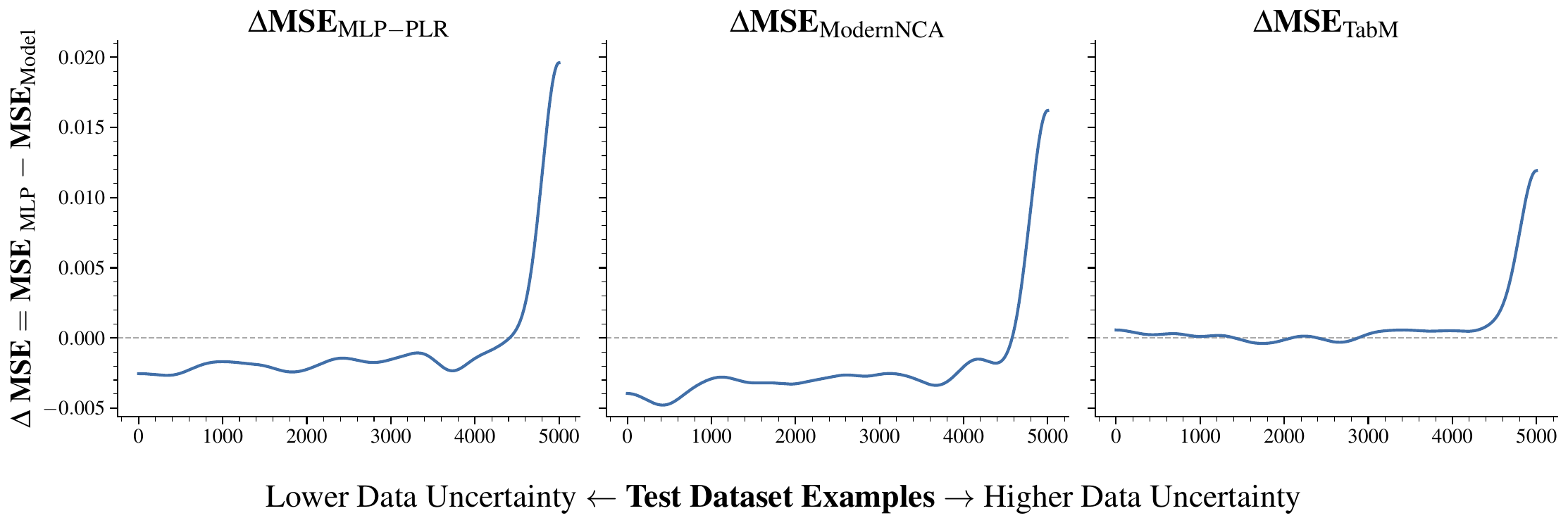}
    \vspace{-5pt}
    \caption{This figure reports performance differences on the synthetic dataset described in this subsection. We report differences in MSE between MLP and each of MLP-PLR, ModernNCA and TabM. Differences are reported for each test example, sorted from left to right by increasing true data uncertainty.}
    \label{fig:synth-compare}
    \vspace{-10pt}
\end{figure*}

First, using this dataset, in \autoref{fig:unc-compare:b} we show that when estimating data uncertainty with CatBoost, the obtained estimates are higly correlated with the true values. Moreover, in \autoref{fig:synth-compare}, we demonstrate that when using true data uncertainty, MLP-PLR, ModernNCA and TabM still show increased performance compared to a simple MLP on high data uncertainty region of the data. 

\subsection{Saw-like 2D synthetic dataset}
\label{sec:saw_spec}

\begin{wrapfigure}{r}{0.45\textwidth}
  \centering
  \vspace{-15pt}
  \includegraphics[width=1\linewidth]{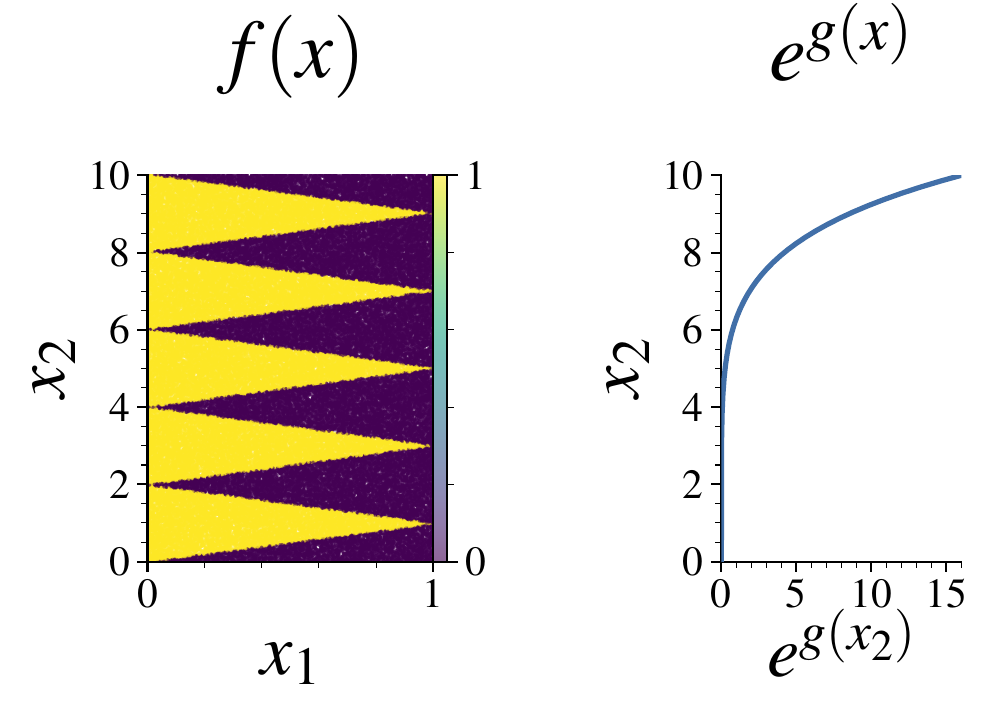} 
  \caption{\hspace{-4pt} The plots show visualization of the synthetic data generation process used in \autoref{sec:saw_spec}. The left plot shows the function $f(x)$ in color, while the right plot shows $e^{g(x)}$ on the $x$-axis. Targets $y$ are generated as $y = f(x) + e^{g(x)} \cdot \mathcal{N}(0, 1)$.}
  \label{fig:saw_generation}
  \vspace{-10pt}
\end{wrapfigure}

In the previous subsection, we demonstrated that the true values of data uncertainty are correlated with increases in the performance of numerical embeddings, ModernNCA, and TabM. As an additional illustration, in this section, we visualize this behavior for the $2D$ dataset presented in \autoref{fig:saw_generation}. 

\begin{figure*}[h!]
    \centering
    \includegraphics[width=\linewidth]{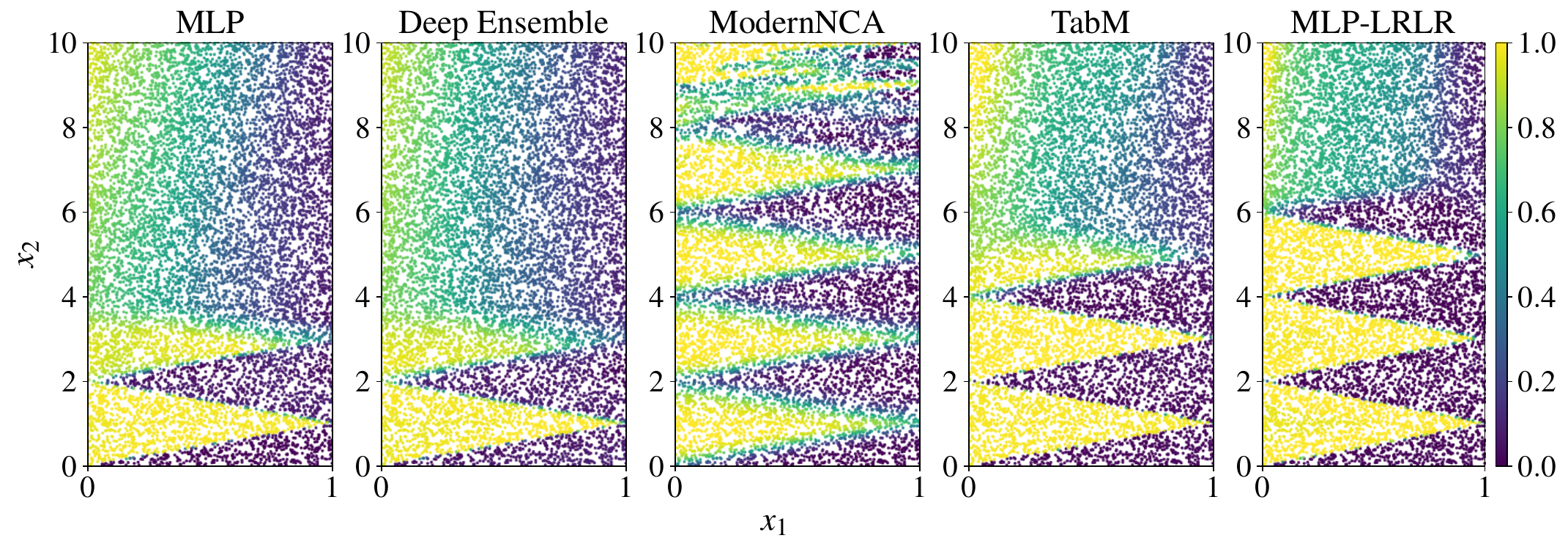}
    \vspace{-15pt}
    \caption{
        The plots show predictions of different models on the synthetic dataset described in \autoref{sec:saw_spec}. While simple ensembling of MLPs does not change the results much, TabM noticeably increases the quality of predictions. ModernNCA shows remarkable ability to recognize the underlying data structure even with high levels of label noise, but it fails to model sharp changes in the target even at low levels of uncertainty. LRLR embeddings remarkably improve the quality of predictions in the regions of high data uncertainty in comparison with a simple MLP.
    }
    \label{fig:saw_predictions}
    \vspace{-10pt}
\end{figure*}

The data were generated as follows: objects $x_i$ are randomly and uniformly sampled from a two-dimensional rectangle with vertices $(0, 0), (1, 0), (1, 10), (0, 10)$. We then construct a saw-like separation line, to the left of which the clean targets $f_i$ are set to 1, and to the right of which the clean targets $f_i$ are 0, as shown in \autoref{fig:saw_generation}. After that step, Gaussian noise with standard deviation $e^{g(x)} = \frac{x_2^6}{62500}$ is added to the targets, where $e^{g(x)}$ depends only on the vertical position of $x$ and grows as its value changes from 0 to 10, as also visualized on \autoref{fig:saw_generation}. The final targets $y_i$ are produced as $y_i = f_i + e^{g(x_i)} \cdot \mathcal{N}(0, 1)$. We use 80,000 training data points and 10,000 validation and test data points each. For the specifics on dataset generation, see \autoref{sec-A:saw-like}.  \autoref{fig:saw_predictions} demonstrates the predictions of the simple MLP model, as well as its following modifications:

\begin{enumerate}
    \item MLP-LRLR --- an MLP model augmented with learned numerical feature embeddings, proposed in \citet{gorishniy2022embeddings}.
    \item ModernNCA \citep{ye2024modern} --- a retrieval-based model that consists of an MLP backbone followed by a kNN-like layer, producing a prediction by aggregating targets of training datapoints.
    \item Deep Ensembles --- the average prediction of five independently trained MLP models.
    \item TabM \citep{gorishniy2024tabm} --- a recent model that imitates an ensemble of MLPs in a parameter-efficient manner.
\end{enumerate}

We provide the exact descriptions of the hyperparameter tuning for each model in \autoref{sec-A:hyper-synth-details}.

From \autoref{fig:saw_predictions}, we see that there are almost no differences in performance between Deep Ensembles and MLP on this dataset. Meanwhile, MLP-LRLR, ModernNCA and TabM show substantially improved performance in the regions of high data uncertainty -- that is, the inner logic of these techniques somehow allows them to manage uncertain regions of the dataset better. Further in the paper, \autoref{sec:embeddings}, \autoref{sec:mnca}, and \autoref{sec:tabm} analyze the mechanisms behind each model in more detail.

\subsection{Synthetic version of the CIFAR-10 dataset with increased data uncertainty}

\begin{wraptable}{r}{0.55\linewidth}
    \centering
    \vspace{-10pt}
    \caption{Accuracy of standard tabular DL methods on two versions of CIFAR-10. MLP outperforms both MLP-LRLR and ModernNCA on the standard CIFAR-10, while TabM and Deep Ensemble perform similarly. On the Uncertain CIFAR-10, both MLP-LRLR and ModernNCA outperform MLP, while TabM outperforms Deep Ensemble. In each category (Single Models and Ensembles), the best method and methods with statistically insignificant differences from it are emphasized in bold.}
    \label{tab:cifar}
    \resizebox{0.9\linewidth}{!}{
\begin{tabular}{lcc}

\multicolumn{3}{c}{Accuracy \textuparrow} \\
\toprule
Method &  CIFAR-10 & Uncertain CIFAR-10 \\
\midrule
\multicolumn{3}{c}{\textbf{Single Model Methods}} \vspace{0.5pt} \\
\midrule
 MLP & \textbf{58.31$\pm$0.27} & 47.54$\pm$0.17 \\
 MLP-LRLR & 57.19$\pm$0.47 & \textbf{48.73$\pm$0.23}\\
 ModernNCA & 57.48$\pm$0.22 & \textbf{48.45$\pm$0.30} \\
\midrule
\multicolumn{3}{c}{\textbf{Emsembling Methods}} \vspace{0.5pt} \\
\midrule
 Deep Ensemble & \textbf{60.92$\pm$0.22} & 49.41$\pm$0.20 \\
 TabM & \textbf{61.01$\pm$0.25} & \textbf{50.45$\pm$0.19} \\
\bottomrule
\vspace{-30pt}
\end{tabular}}
\end{wraptable}

In this section, we construct another synthetic dataset based on the established CIFAR-10 dataset \citep{krizhevsky2009learning}. To do this, we use the intuition that data uncertainty may increase when some features relevant to the target variable are omitted from the dataset. Specifically, we take a standard version of CIFAR-10, which has $32 \times 32 \times 3 = 3072$ features, and then randomly choose $50$ features to keep, dropping all the other features from each sample in the dataset. As a result of this operation, we obtain a dataset we refer to as Uncertain CIFAR-10.

We then conduct experiments on both the pruned and the original versions of CIFAR-10, using standard tabular DL methods. As can be seen from \autoref{tab:cifar}, on the original version of the dataset, MLP outperforms both other single model methods (MLP-LRLR and ModernNCA), and in ensembling methods there is no difference between standard Deep Ensemble and TabM. However, on the Uncertain CIFAR-10, TabM outperforms Deep Ensemble, while MLP-LRLR and ModernNCA outperform MLP. This experiment provides additional evidence of a relationship between data uncertainty and increases in performance for different tabular methods.

\section{Embeddings for numerical features}
\label{sec:embeddings}

Recent work by \citet{gorishniy2022embeddings} demonstrated that, for tabular DL, embedding scalar numerical features into a high-dimensional space before they are fed into the main backbone is beneficial. While numerical embeddings have become an established design choice in the field, the reasons for their effectiveness are not entirely clear. In our paper, we reveal that an important property of numerical embeddings is their ability to handle high data uncertainty regions of data.

\autoref{fig:plr-compare} demonstrates uncertainty plots for MLP vs MLP-PLR, where ``-PLR'' denotes periodic activations followed by a linear layer with a ReLU nonlinearity as described in \citet{gorishniy2022embeddings}. Here we report the analysis on four datasets\footnote{Similar plots on other established datasets are reported in \autoref{sec-A:full-uncertainty}.}:

\begin{enumerate}
\item California Housing\footnote{\url{https://www.dcc.fc.up.pt/~ltorgo/Regression/cal_housing.html}}, an established regression dataset, where numerical embeddings were shown to be beneficial \citep{gorishniy2022embeddings}.

\item Sberbank Housing \citep{rubachev2024tabred}, a recently introduced dataset with a temporal distribution shift between its training and test subsets.

\item Maps Routing \citep{rubachev2024tabred}, a large-scale dataset needed to demonstrate that our analysis is valid for larger problems as well.

\item Delivery ETA \citep{rubachev2024tabred}, a dataset where MLP-PLR underperforms MLP. Despite this, the model with embeddings still slightly outperforms the MLP baseline in the high data uncertainty region, while losing more in the lower data uncertainty niche. 

\begin{figure*}[h!]
    \centering
    \includegraphics[width=\linewidth]{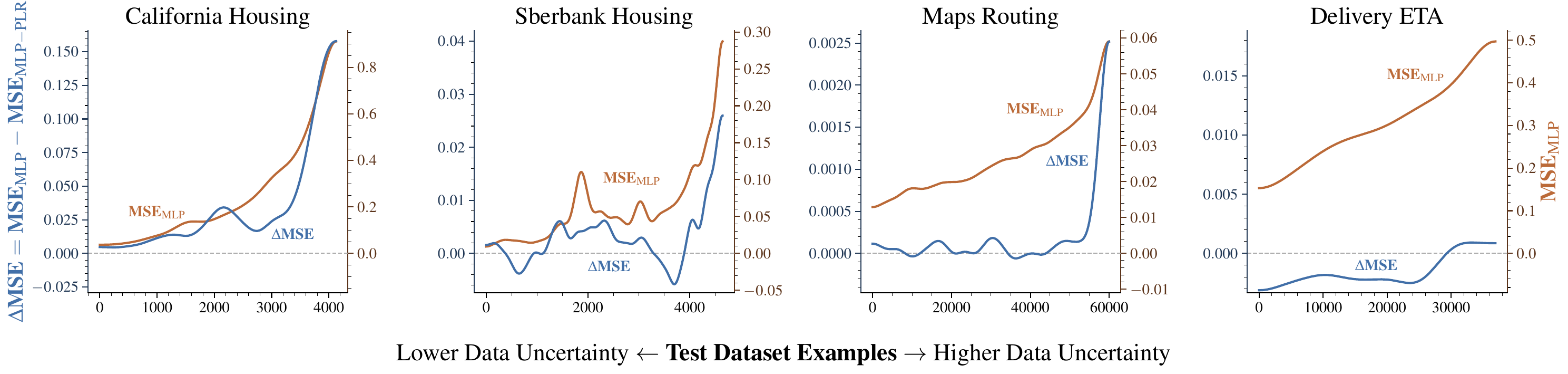}
    \caption{Uncertainty plots report the differences in MSE between MLP and MLP-PLR. The increase in performance is the most significant in the high data uncertainty niche.}
    \label{fig:plr-compare}
    \vspace{-5pt}
\end{figure*}
\end{enumerate}

The main observation from \autoref{fig:plr-compare} is that numerical ``-PLR'' embeddings are disproportionately more beneficial for high data uncertainty samples. Specifically, the difference in MSE grows faster with data uncertainty than the value of MSE for the baseline MLP, which shows that numerical embeddings do not just proportionally improve performance for all levels of uncertainty, but do so more in high data uncertainty region.

\subsection{Why do numerical embeddings outperform in the high data uncertainty niche?}

\begin{figure*}[h!]
    \centering
    \includegraphics[width=0.95\linewidth]{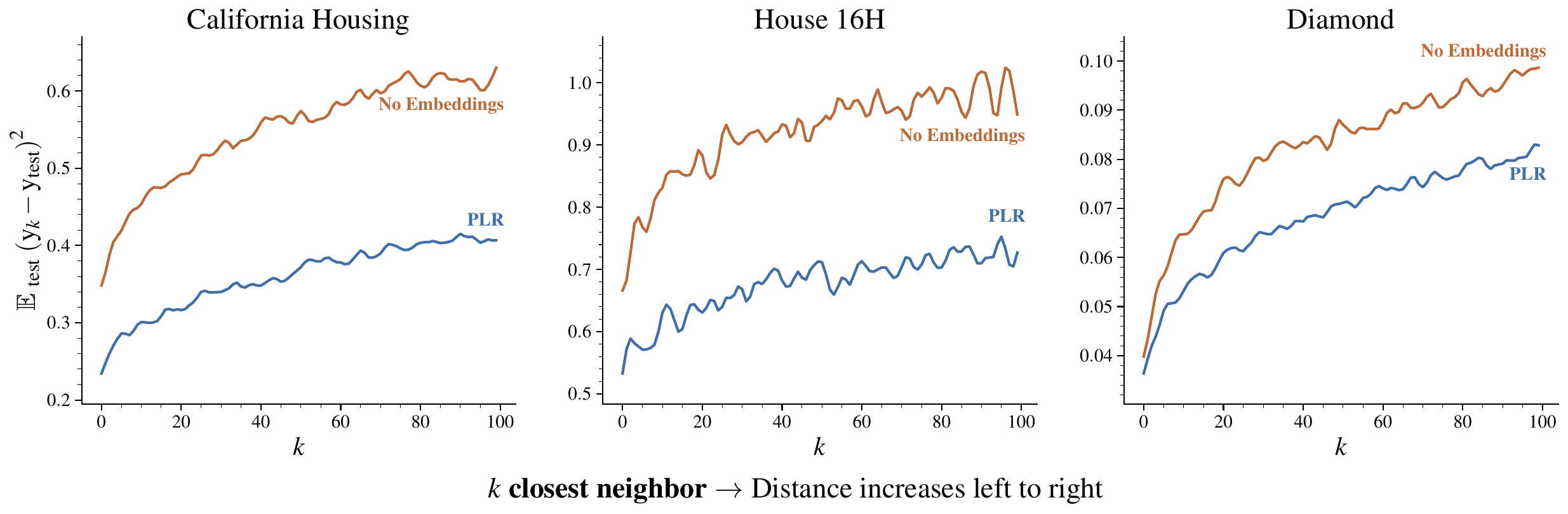}
    \caption{Squared difference between a test sample's target and the target of its $k$-th closest neighbor, averaged over the test set. The difference is smaller when using PLR embeddings.}
    \label{fig:emb_y_degradation}
    \vspace{-15pt}
\end{figure*}

When using numerical embeddings, one maps the original datapoints into a higher-dimensional space that effectively becomes a new input space for the main backbone. We claim that properly designed numerical embeddings correspond to a space with higher local target consistency, i.e., the neighboring points in this space are more likely to have similar targets. We demonstrate that behavior on several datasets. \autoref{fig:emb_y_degradation} reports the squared difference between the target of a test datapoint and the target of its $k$-th closest neighbor from the training set (averaged over all test subset) for MLP and MLP-PLR models. We sort the neighbors by the Euclidean distance in the latent space obtained after the first block of the model, to show the distance in the space that has already incorporated the signal from all features and still comes relatively early in the model. 

While numerical embeddings improve the local target consistency in the neighborhoods of all datapoints, this leads to disproportionate performance improvement on datapoints that have higher data uncertainty, as shown in \autoref{fig:plr-compare}. We explain this effect by the fact that the model's predictions on high data uncertainty points rely on the quality of their neighborhoods more heavily. \autoref{fig:saw_predictions} confirms our intuition on the synthetic task. For low data uncertainty regions (bottom), a simple MLP is able to model sharp separating planes between the predictions ``$0$'' and ``$1$''. As data uncertainty increases, the MLP predictions become smoother, until they become so smooth that the original saw-like target structure becomes unnoticeable. This demonstrates that when MLP makes a prediction on a high data uncertainty datapoint, it is more dependent on the datapoint's neighborhood, effectively performing more ``conservative'' local target smoothing. As an additional illustration, \autoref{fig:saw_predictions} shows that the better representations provided by the learned LRLR embeddings lead to better predictions even at the levels where the original MLP predictions fail. 

\subsection{A more effective embedding scheme}

\begin{wraptable}{r}{0.55\linewidth}
    \centering
    \caption{This table shows a comparison of the tuned MLP, the tuned MLP with LRLR embeddings and the tuned MLP with our proposed improved LRLR embeddings trained to maximize local target consistency. We report average percentage improvements over MLP and average ranks across regression datasets from \citet{grinsztajn2022why, gorishniy2024tabm}. We can see that the LRLR\textsubscript{triplet} embeddings substantially outperform the existing alternatives. For full dataset-wise results, see \autoref{sec-A:scheme}}
    \label{tab:emb_pretrain}
    \resizebox{0.8\linewidth}{!}{
    \begin{tabular}{llc}
        \toprule
        Model & $\Delta$ MLP (\%) & Avg. Rank \\
        \midrule
        MLP                        & 0.00    & 2.75 \textsubscript{(\textpm 0.50)} 	 \\
        MLP-LRLR                   & 2.00 \textsubscript{(\textpm 2.34)}   & 1.67 \textsubscript{(\textpm 0.49)} \\
        \midrule
        MLP-LRLR\textsubscript{triplet} & 2.81 \textsubscript{(\textpm 2.54)}   & 1.17 \textsubscript{(\textpm 0.51)} \\
        \bottomrule
    \end{tabular}
    }
\end{wraptable}

While numerical embeddings proposed in \citet{gorishniy2022embeddings} substantially improve the quality of neighborhoods needed to handle high data uncertainty, they do not explicitly aim for this behavior. In contrast, we propose a new numerical embedding scheme that is learned by explicitly maximizing the local target consistency. In more detail, we train an embedder module that produces a high-dimensional representation of a given object. The architecture of our module consists of the ``-LRLR'' embedder described in \citet{gorishniy2022embeddings}, followed by a simple linear layer.

We train the embedder module using the standard triplet loss \citep{Schroff_2015_CVPR}. More specifically, we first randomly sample a batch of training objects that will serve as anchors in the triplet loss. Then, for each anchor object, we randomly sample a pair of other training objects, of which the one with the closer target becomes a positive example, and the other one becomes the negative example. We then compute the embeddings for all three objects with our embedder as well as two dot-product similarities: the first one is between the embeddings of the anchor and the positive example, and the second one is between the embeddings of anchor and the negative example. We then pass these similarities as logits to the cross-entropy loss, for which the target is ``0''. By doing this, we effectively force the model to learn closer representations for objects with more similar targets. 

After pretraining the embedder, we discard its last linear layer and use its remaining part to initialize the embedding part of the ``MLP-LRLR'' architecture. The exact hyperparameters used in our experiments are provided in \autoref{sec-A:scheme}. 

\autoref{tab:emb_pretrain} demonstrates how our embedding scheme improves the performance in comparison with LRLR embeddings from \citet{gorishniy2022embeddings} trained from scratch. Information regarding statistical significance, along with full results for each dataset, is provided in \autoref{sec-A:scheme}.

\section{ModernNCA and high data uncertainty}
\label{sec:mnca}

\begin{figure*}[h!]
    \centering
    \includegraphics[width=\linewidth]{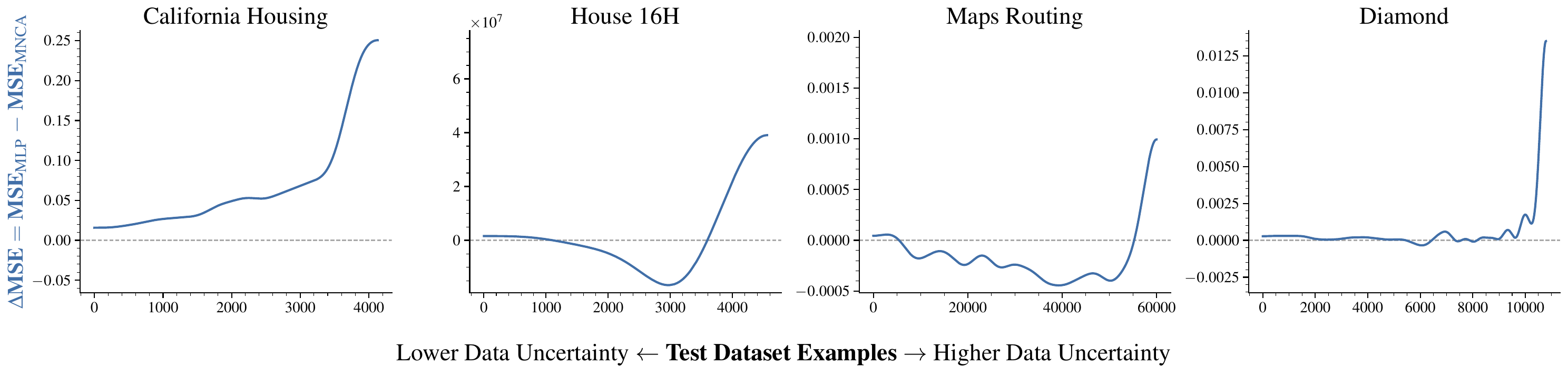}
    \caption{Uncertainty plots, built in the same way as in \autoref{fig:xgb-compare}, but showing difference in MSE between MLP and ModernNCA. On House and Maps-Routing datasets, ModernNCA is inferior to MLP in the lower/medium data uncertainty niches, while on the other two datasets it shows improvement in all niches of uncertainty.}
    \label{fig:mnca-compare}
\end{figure*}

To investigate the interplay between retrieval modules and data uncertainty, we focus on the ModernNCA model \citep{ye2024modern}, which is a recent model that consists of an MLP-like backbone followed by a kNN-like prediction head. \autoref{fig:mnca-compare} shows that compared to MLP, ModernNCA also improves the performance on higher data uncertainty points, while the performance on lower data uncertainty points sometimes slightly decreases. We explain this by first noticing that averaging over a large number of neighbors inherent to retrieval-augmented models leads to less overfitting on the high data uncertainty regions when compared to MLP. 

\begin{figure*}[h!]
    \centering
    \includegraphics[width=\linewidth]{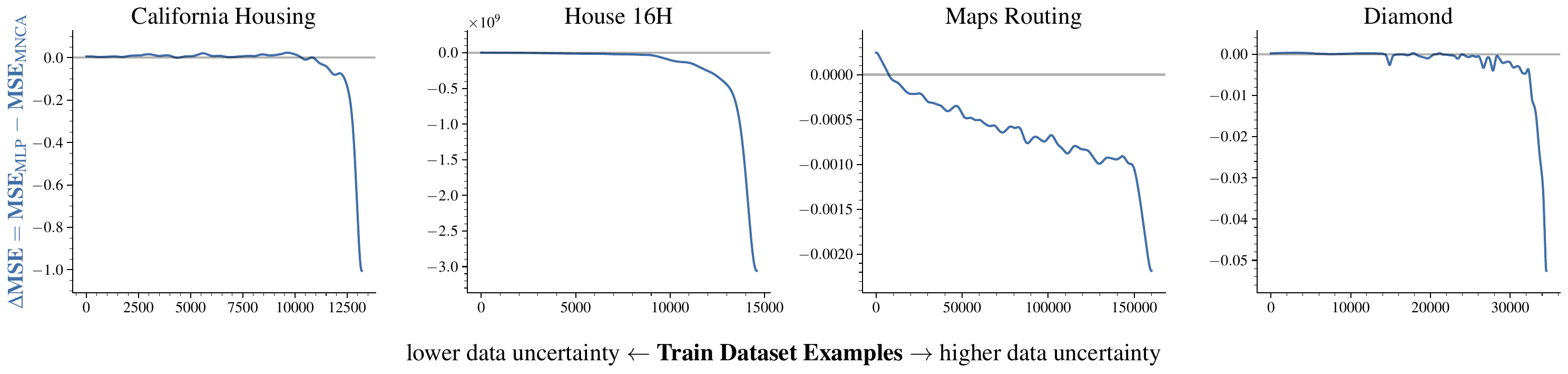}
    \caption{Difference in MSE performance of MLP and ModernNCA on train samples, with samples sorted by increasing uncertainty. ModernNCA's higher loss demonstrates that it often produces overly smooth predictions, underfitting the data, for example in \autoref{fig:saw_predictions}.}
    \label{fig:mnca-train-compare}
\end{figure*}

To demonstrate this, in \autoref{fig:mnca-train-compare} we report the sample-wise MSE loss of ModernNCA computed on training datapoints, where training datapoints are sorted in order of the increasing data uncertainty. \autoref{fig:mnca-train-compare} shows that compared to MLP, ModernNCA severely undefits the data on high data uncertainty zone. On the one hand, this underfitting effect provides robustness to the noise in the train targets for the high data uncertainty niche, since ModernNCA memorizes the noise in the data to a lesser degree. On the other hand, on some datasets (e.g. Maps Routing, House), this excessive underfit can be harmful to performance in the lower and medium data uncertainty niches. 

\section{Ensembling and high data uncertainty}
\label{sec:tabm}
\begin{figure*}[h!]
    \centering
    \includegraphics[width=\linewidth]{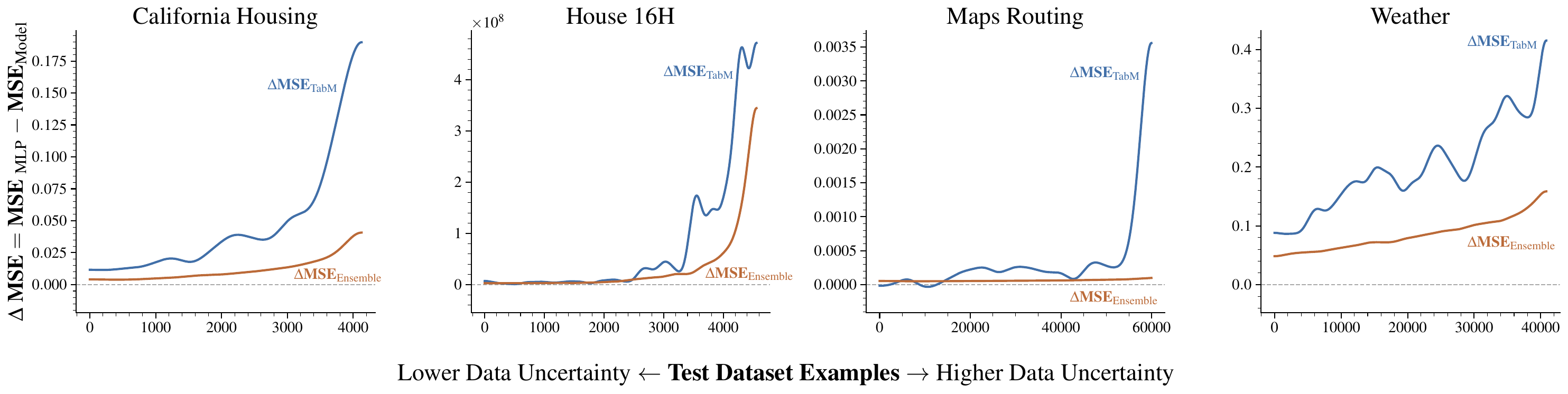}
    \caption{The plots show sample-wise differences in MSE between MLP and TabM in blue and sample-wise differences in MSE between MLP and a Deep Ensemble of MLPs in brown, with samples sorted by increasing data uncertainty. TabM shows substantially higher performance, especially on high data uncertainty regions.}
    \label{fig:tabm_compare}
\end{figure*}

The recent TabM model \citep{gorishniy2024tabm} also demonstrates a more significant improvement on higher data uncertainty datapoints, as shown in \autoref{fig:tabm_compare}. Interestingly, while TabM was described as an efficient ensemble, traditional deep ensemble demonstrates inferior performance, which is especially noticeable in high data uncertainty regions. Another illustration of the differences between TabM and ensemble is shown by our synthetic example described in \autoref{sec:saw_spec}. While deep ensemble shows almost no difference in performance compared to a single MLP, TabM learns a function much better representing the underlying noiseless target distribution.

In this section, we explain the reason behind the robustness of TabM to noise in the targets compared to a deep ensemble. Below, we demonstrate that this stems from the fact that the averaging of the gradients for shared parameters of TabM effectively reduces the noise in the gradients from individual branches.

\begin{figure}[h!]
    \begin{minipage}{0.48\textwidth}
        \centering
        \vspace{-10pt}
          \includegraphics[width=1\linewidth]{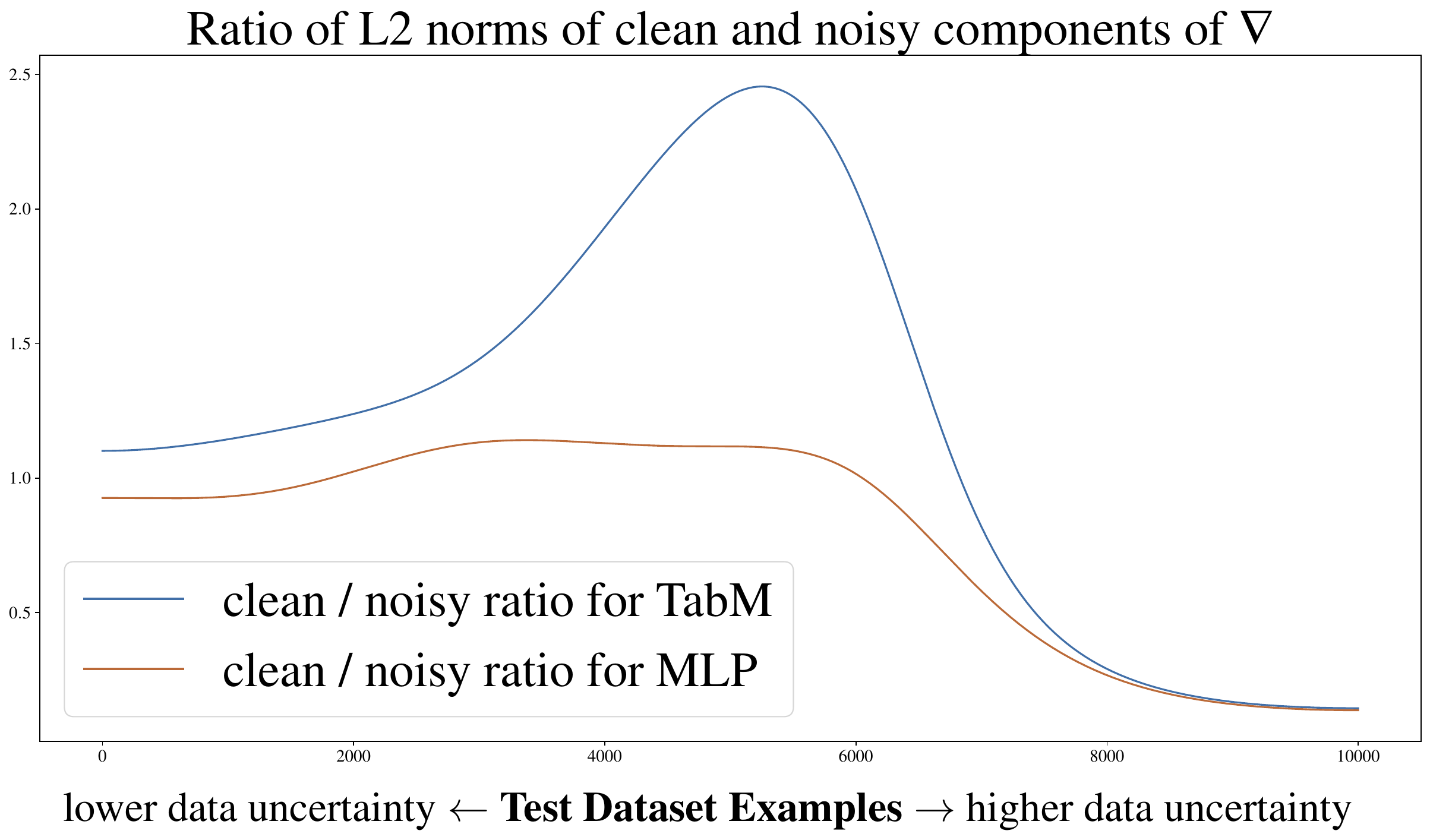} 
          \caption{The plot shows ratios of the norms of clean and noisy components of the gradient for TabM and MLP for the synthetic dataset from \autoref{sec:saw_spec}. In this experiment, both models were tuned and trained. Data uncertainty increases from left to right. The ratio grows with increasing data uncertainty more for TabM until it reaches a point in data uncertainty where both models fail; after that it starts moving towards the value of MLP's clean-to-noisy ratio.}
          \label{fig:grad_ratios}
            \end{minipage}\hfill
    \begin{minipage}{0.48\textwidth}
        \centering
        \includegraphics[width=1\linewidth]{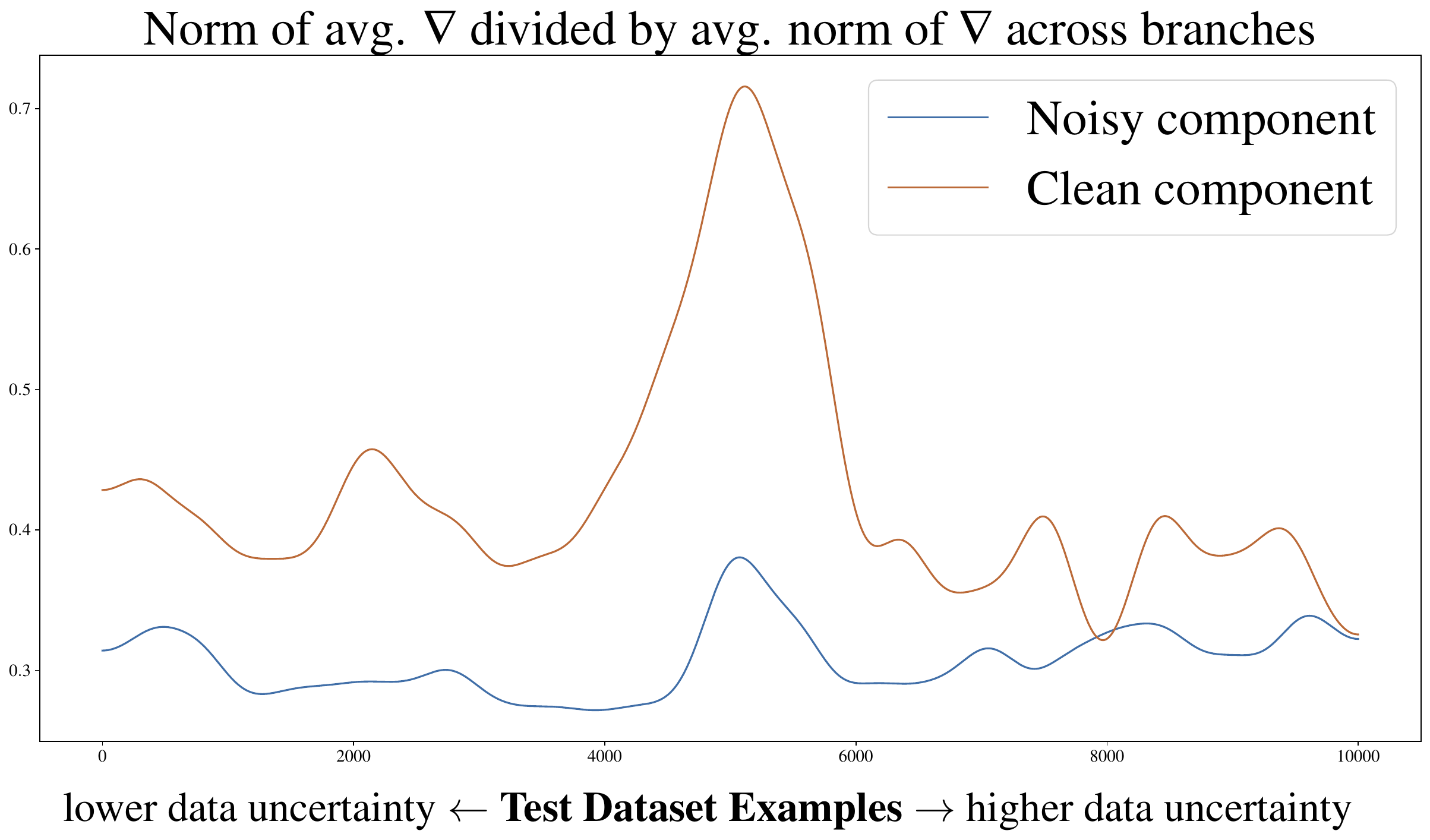} 
        \caption{The plot shows the ratio of the average norm of a per-branch gradient to a norm of the gradient averaged between branches, taken separately for clean and noisy components. In this experiment, TabM model was tuned and trained. Higher value shows alignment between the individual gradients. The ratio grows with increasing data uncertainty more for the clean component, until it reaches a point in data uncertainty where TabM fails, after which it approaches the ratio for the noisy part of the gradients.}
        \label{fig:consistency}
    \end{minipage}
\end{figure}

First, let us decompose the model's gradient for a particular datapoint into a ``clean'' component and a ``noisy'' component. We denote the model as a function $\phi(\cdot)$ and denote the model weights as $\theta$. As described in \autoref{sec:saw_spec}, the target values $y_i$ are comprised of the noiseless part and the noisy term $y_i = f(x_i) + e^{g(x_i)} \cdot \mathcal{N}(0, 1)$. Therefore,

\begin{equation}
    \begin{split}
& \frac{\partial(\phi(x_i) - y_i)^2}{\partial\theta} = 2(\phi(x_i) - y_i)\frac{\partial\phi(x_i)}{\partial\theta} = 2(\phi(x_i) - f(x_i) - e^{g(x_i)} \cdot \mathcal{N}(0, 1)) \frac{\partial\phi(x_i)}{\partial\theta} = \\
&= 2(\phi(x_i) - f(x_i)) \frac{\partial\phi(x_i)}{\partial\theta} - 2 e^{g(x_i)} \cdot \mathcal{N}(0, 1) \cdot \frac{\partial\phi(x_i)}{\partial\theta}
\end{split}
\end{equation}

In this subtraction, the first term 
corresponds to the gradient of the noiseless part of 
the target dependency $f(\cdot)$. The second term corresponds to the noisy component of the gradient. Both terms can be explicitly computed for the synthetic data from \autoref{sec:saw_spec}, since both $f(\cdot)$ and $g(\cdot)$ are known.

For robust training on noisy data, it is beneficial to minimize the noisy component of the gradients to make overall gradients as close to their noiseless components as possible. We argue that the contribution from the noisy components to the overall gradients is much lower in TabM compared to MLP. To illustrate this, in \autoref{fig:grad_ratios}, we show that the ratio of L2 norms of clean and noisy gradient components is much higher in TabM than in MLP, indicating that the noisy gradients affect TabM's learning process to a lesser extent.

We explain this difference by averaging of the gradients obtained from different branches in TabM. To confirm this explanation, we compute the ratio of the norm of averaged gradients to the average norm of gradient of the individual branches (for both the noisy and clean gradient components).
In \autoref{fig:consistency}, we report this ratio and show that the averaging reduces the norm of the noisy part more substantially than that of the clean part.

This behavior makes TabM more reluctant to learn noisy parts of the target dependency, which leads to better performance in the high data uncertainty region.

\section{Conclusion}

In this paper, we propose an uncertainty-centric lens to investigate tabular DL approaches. This unifying perspective serves a dual purpose: it elucidates the strengths and limitations of various recent techniques and directly informs the development of a more effective scheme for embedding numerical features. Looking ahead, an interesting avenue for future research involves extending this uncertainty-driven analysis to the growing field of tabular foundational models \citep{hollmann2025accurate}.

\section{Limitations}

Several limitations should be acknowledged in this work. Our investigation was primarily focused on regression tasks; future research could extend this analysis to include classification problems. Furthermore, to maintain a specific focus on tabular DL approaches, we deliberately excluded the examination of general-purpose deep learning mechanisms (e.g., dropout, weight decay, robust optimizers). Lastly, although knowledge uncertainty presents an important and potentially informative viewpoint, its detailed exploration was not a central objective of the current study.

\bibliographystyle{refstyle}
\bibliography{references}

\begin{thebibliography}{27}
\providecommand{\natexlab}[1]{#1}
\providecommand{\url}[1]{\texttt{#1}}
\expandafter\ifx\csname urlstyle\endcsname\relax
  \providecommand{\doi}[1]{doi: #1}\else
  \providecommand{\doi}{doi: \begingroup \urlstyle{rm}\Url}\fi

\bibitem[Baek et~al.(2024)Baek, Kolter, and Raghunathan]{baek2024sam}
Baek, C., Kolter, Z., and Raghunathan, A.
\newblock Why is sam robust to label noise?
\newblock In \emph{ICLR}, 2024.

\bibitem[Bahri et~al.(2021)Bahri, Jiang, Tay, and Metzler]{bahri2021scarf}
Bahri, D., Jiang, H., Tay, Y., and Metzler, D.
\newblock {SCARF}: Self-supervised contrastive learning using random feature corruption.
\newblock In \emph{ICLR}, 2021.

\bibitem[Chen \& Guestrin(2016)Chen and Guestrin]{chen2016xgboost}
Chen, T. and Guestrin, C.
\newblock {XGB}oost: A scalable tree boosting system.
\newblock In \emph{SIGKDD}, 2016.

\bibitem[Duan et~al.(2020)Duan, Anand, Ding, Thai, Basu, Ng, and Schuler]{duan2020ngboost}
Duan, T., Anand, A., Ding, D.~Y., Thai, K.~K., Basu, S., Ng, A., and Schuler, A.
\newblock Ngboost: Natural gradient boosting for probabilistic prediction.
\newblock In \emph{International conference on machine learning}, pp.\  2690--2700. PMLR, 2020.

\bibitem[Gal et~al.(2016)]{gal2016uncertainty}
Gal, Y. et~al.
\newblock Uncertainty in deep learning.
\newblock \emph{phd thesis, University of Cambridge}, 2016.

\bibitem[Gorishniy et~al.(2021)Gorishniy, Rubachev, Khrulkov, and Babenko]{gorishniy2021revisiting}
Gorishniy, Y., Rubachev, I., Khrulkov, V., and Babenko, A.
\newblock Revisiting deep learning models for tabular data.
\newblock In \emph{NeurIPS}, 2021.

\bibitem[Gorishniy et~al.(2022)Gorishniy, Rubachev, and Babenko]{gorishniy2022embeddings}
Gorishniy, Y., Rubachev, I., and Babenko, A.
\newblock On embeddings for numerical features in tabular deep learning.
\newblock In \emph{NeurIPS}, 2022.

\bibitem[Gorishniy et~al.(2024)Gorishniy, Rubachev, Kartashev, Shlenskii, Kotelnikov, and Babenko]{gorishniy2023tabr}
Gorishniy, Y., Rubachev, I., Kartashev, N., Shlenskii, D., Kotelnikov, A., and Babenko, A.
\newblock Tab{R}: Tabular deep learning meets nearest neighbors.
\newblock In \emph{ICLR}, 2024.

\bibitem[Gorishniy et~al.(2025)Gorishniy, Kotelnikov, and Babenko]{gorishniy2024tabm}
Gorishniy, Y., Kotelnikov, A., and Babenko, A.
\newblock Tabm: Advancing tabular deep learning with parameter-efficient ensembling.
\newblock In \emph{ICLR}, 2025.

\bibitem[Grinsztajn et~al.(2022)Grinsztajn, Oyallon, and Varoquaux]{grinsztajn2022why}
Grinsztajn, L., Oyallon, E., and Varoquaux, G.
\newblock Why do tree-based models still outperform deep learning on typical tabular data?
\newblock In \emph{NeurIPS, the "Datasets and Benchmarks" track}, 2022.

\bibitem[Hollmann et~al.(2023)Hollmann, Müller, Eggensperger, and Hutter]{hollmann2022tabpfn}
Hollmann, N., Müller, S., Eggensperger, K., and Hutter, F.
\newblock Tab{PFN}: A transformer that solves small tabular classification problems in a second.
\newblock In \emph{ICLR}, 2023.

\bibitem[Hollmann et~al.(2025)Hollmann, M{\"u}ller, Purucker, Krishnakumar, K{\"o}rfer, Hoo, Schirrmeister, and Hutter]{hollmann2025accurate}
Hollmann, N., M{\"u}ller, S., Purucker, L., Krishnakumar, A., K{\"o}rfer, M., Hoo, S.~B., Schirrmeister, R.~T., and Hutter, F.
\newblock Accurate predictions on small data with a tabular foundation model.
\newblock \emph{Nature}, 637\penalty0 (8045):\penalty0 319--326, 2025.

\bibitem[Holzm{\"u}ller et~al.(2024)Holzm{\"u}ller, Grinsztajn, and Steinwart]{holzmüller2024better}
Holzm{\"u}ller, D., Grinsztajn, L., and Steinwart, I.
\newblock Better by default: Strong pre-tuned mlps and boosted trees on tabular data.
\newblock In \emph{The Thirty-eighth Annual Conference on Neural Information Processing Systems}, 2024.

\bibitem[Jeffares et~al.(2023)Jeffares, Liu, Crabbé, Imrie, and van~der Schaar]{jeffares2023tangos}
Jeffares, A., Liu, T., Crabbé, J., Imrie, F., and van~der Schaar, M.
\newblock {TANGOS}: Regularizing tabular neural networks through gradient orthogonalization and specialization.
\newblock In \emph{ICLR}, 2023.

\bibitem[Ke et~al.(2017)Ke, Meng, Finley, Wang, Chen, Ma, Ye, and Liu]{ke2017lightgbm}
Ke, G., Meng, Q., Finley, T., Wang, T., Chen, W., Ma, W., Ye, Q., and Liu, T.-Y.
\newblock Light{GBM}: A highly efficient gradient boosting decision tree.
\newblock \emph{Advances in neural information processing systems}, 30:\penalty0 3146--3154, 2017.

\bibitem[Krizhevsky et~al.(2009)Krizhevsky, Hinton, et~al.]{krizhevsky2009learning}
Krizhevsky, A., Hinton, G., et~al.
\newblock Learning multiple layers of features from tiny images.(2009), 2009.

\bibitem[Malinin et~al.(2021)Malinin, Prokhorenkova, and Ustimenko]{malinin2020uncertainty}
Malinin, A., Prokhorenkova, L., and Ustimenko, A.
\newblock Uncertainty in gradient boosting via ensembles.
\newblock In \emph{ICLR}, 2021.

\bibitem[McElfresh et~al.(2023)McElfresh, Khandagale, Valverde, Prasad~C, Ramakrishnan, Goldblum, and White]{mcelfresh2023neural}
McElfresh, D., Khandagale, S., Valverde, J., Prasad~C, V., Ramakrishnan, G., Goldblum, M., and White, C.
\newblock When do neural nets outperform boosted trees on tabular data?
\newblock \emph{Advances in Neural Information Processing Systems}, 36:\penalty0 76336--76369, 2023.

\bibitem[Prokhorenkova et~al.(2018)Prokhorenkova, Gusev, Vorobev, Dorogush, and Gulin]{prokhorenkova2018catboost}
Prokhorenkova, L., Gusev, G., Vorobev, A., Dorogush, A.~V., and Gulin, A.
\newblock Cat{B}oost: unbiased boosting with categorical features.
\newblock In \emph{NeurIPS}, 2018.

\bibitem[Rubachev et~al.(2022)Rubachev, Alekberov, Gorishniy, and Babenko]{rubachev2022revisiting}
Rubachev, I., Alekberov, A., Gorishniy, Y., and Babenko, A.
\newblock Revisiting pretraining objectives for tabular deep learning.
\newblock \emph{arXiv}, 2207.03208v1, 2022.

\bibitem[Rubachev et~al.(2025)Rubachev, Kartashev, Gorishniy, and Babenko]{rubachev2024tabred}
Rubachev, I., Kartashev, N., Gorishniy, Y., and Babenko, A.
\newblock {TabReD: Analyzing Pitfalls and Filling the Gaps in Tabular Deep Learning Benchmarks}.
\newblock In \emph{arXiv}, 2025.

\bibitem[Schroff et~al.(2015)Schroff, Kalenichenko, and Philbin]{Schroff_2015_CVPR}
Schroff, F., Kalenichenko, D., and Philbin, J.
\newblock Facenet: A unified embedding for face recognition and clustering.
\newblock In \emph{Proceedings of the IEEE Conference on Computer Vision and Pattern Recognition (CVPR)}, June 2015.

\bibitem[Somepalli et~al.(2021)Somepalli, Goldblum, Schwarzschild, Bruss, and Goldstein]{somepalli2021saint}
Somepalli, G., Goldblum, M., Schwarzschild, A., Bruss, C.~B., and Goldstein, T.
\newblock {SAINT:} improved neural networks for tabular data via row attention and contrastive pre-training.
\newblock \emph{arXiv}, 2106.01342v1, 2021.

\bibitem[Tanaka et~al.(2018)Tanaka, Ikami, Yamasaki, and Aizawa]{tanaka2018jointoptimizationframeworklearning}
Tanaka, D., Ikami, D., Yamasaki, T., and Aizawa, K.
\newblock Joint optimization framework for learning with noisy labels, 2018.
\newblock URL \url{https://arxiv.org/abs/1803.11364}.

\bibitem[Thimonier et~al.(2024)Thimonier, Costa, Popineau, Rimmel, and Doan]{thimonier2024t}
Thimonier, H., Costa, J. L. D.~M., Popineau, F., Rimmel, A., and Doan, B.-L.
\newblock T-jepa: Augmentation-free self-supervised learning for tabular data.
\newblock In \emph{ICLR}, 2024.

\bibitem[Tschalzev et~al.(2025)Tschalzev, Purucker, L{\"u}dtke, Hutter, Bartelt, and Stuckenschmidt]{tschalzev2025unreflected}
Tschalzev, A., Purucker, L., L{\"u}dtke, S., Hutter, F., Bartelt, C., and Stuckenschmidt, H.
\newblock Unreflected use of tabular data repositories can undermine research quality.
\newblock \emph{arXiv preprint arXiv:2503.09159}, 2025.

\bibitem[Ye et~al.(2024)Ye, Yin, and Zhan]{ye2024modern}
Ye, H.-J., Yin, H.-H., and Zhan, D.-C.
\newblock Modern neighborhood components analysis: A deep tabular baseline two decades later.
\newblock \emph{arXiv}, 2407.03257v1, 2024.

\end{thebibliography}



\newpage

\appendix

\section{Uncertainty plots for other datasets}
\label{sec-A:full-uncertainty}

\begin{figure*}[h!]
    \centering
    \includegraphics[width=\linewidth]{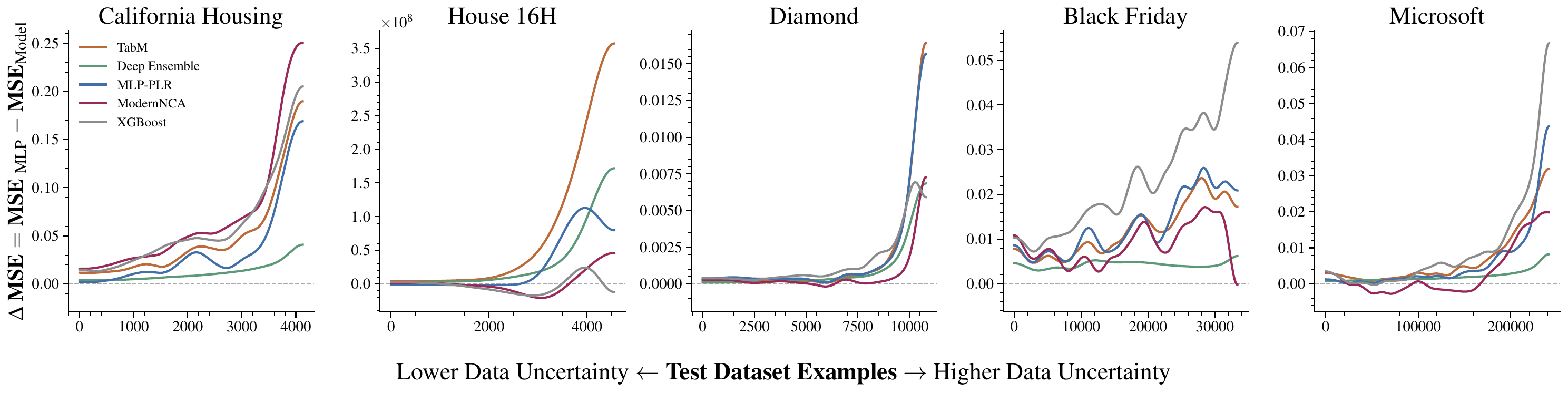}
    \caption{Uncertainty plots depicting differences between MLP and XGBoost, Deep Ensemble, MLP-PLR, ModernNCA and TabM.}
    \label{fig-A:full-default}
\end{figure*}

\begin{figure*}[h!]
    \centering
    \includegraphics[width=\linewidth]{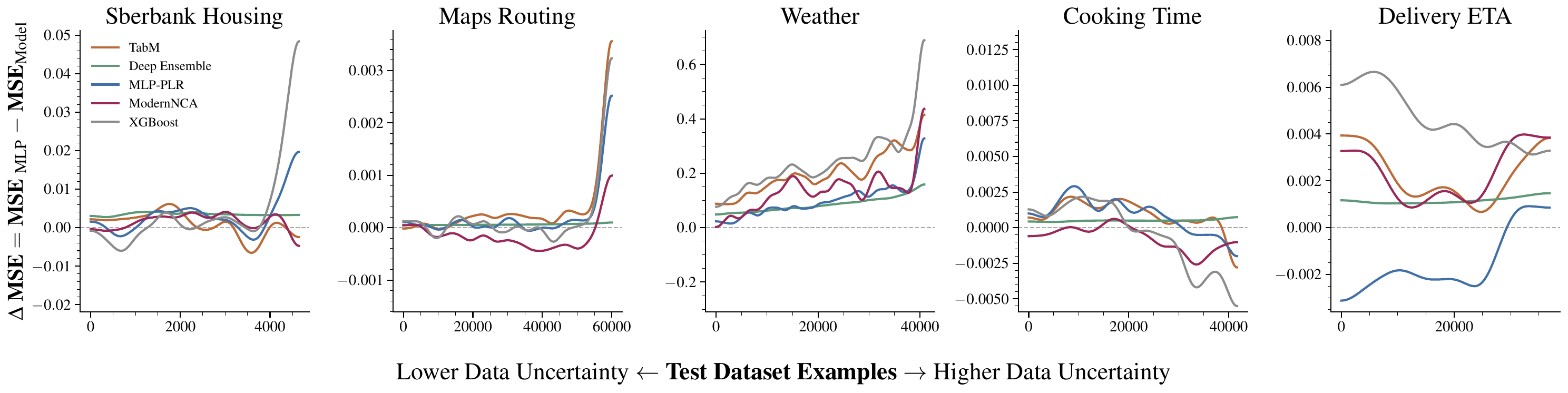}
    \caption{Uncertainty plots depicting differences between MLP and XGBoost, Deep Ensemble, MLP-PLR, ModernNCA and TabM. The Cooking Time dataset is an exception to our analysis, which we believe stems from the failure of data uncertainty estimation on this dataset.}
    \label{fig-A:full-tabred}
\end{figure*}

In this section we provide the uncertainty plots for all regression datasets from the \citet{rubachev2024tabred} and \citet{gorishniy2021revisiting} benchmarks. In the main text we report the most informative plots that clearly illustrate our observations. The uncertainty plots for all models on all datasets are in \autoref{fig-A:full-default} and \autoref{fig-A:full-tabred}.

\section{Technical details regarding uncertainty plots}
\label{sec-A:uncertainty-details}

To estimate data uncertainty, we use CatBoost with the RMSEWithUncertainty loss function. Then, we sort the test datapoints in ascending order based on their estimated data uncertainty values, and calculate the difference in MSE between two models for each datapoint. Finally, we apply gaussian smoothing, using function gaussial\_filter1d from scipy library.

\section{Technical details regarding hyperparameter tuning for models in empirical experiments}
\label{sec-A:hyper-emp-details}

In this section we describe the hyperparameter spaces that were used to tune model hyperparameters in the experiments.

For the TabM model, we use the default TabM model from the reference implementation and follow \citet{gorishniy2024tabm} tuning spaces. For ModernNCA, MLP and MLP-PLR we also follow the tuning spaces from \citet{gorishniy2024tabm}, except using powers of two for the neural network width tuning. For the exact tuning spaces and best parameters see the source code provided with the submission.

\section{Technical details regarding a synthetic saw-like dataset}
\label{sec-A:saw-like}

The exact algorithm used for the generation of the dataset is provided in the code. Generally speaking, we uniformly sample points from (0, 0) to (1, 10), and then set the target value as 1 inside the triangles with vertices $(2i, 0), (2i + 1, 1), (2i + 2, 0)$, for values of i: $0, 1, 2, 3, 4$. We also add the random normal noise to the targets, with the standard deviation of the noise set to $\frac{x_1^6}{4}$, for $x_1$ from 0 to 10. 

\section{Technical details regarding hyperparameter tuning for models on synthetic datasets}
\label{sec-A:hyper-synth-details}

Here we describe the technical details regarding hyperparameter tuning for synthetic datasets.

\textbf{Saw-like synthetic data}. For this experiment we keep all model-architecture hyperparameters fixed at reasonable default values (3 layers, hidden dimension is 256, dropout is 0.2, 16 branches for TabM, dimension of embedding is 64 for MLP-LRLR) and only tune the learning rates, except the ModernNCA model for which we also tune the architecture hyperparameters. Details regarding tuning, early-stopping and other training protocols follow the one described in \autoref{sec-A:hyper-emp-details}.

\textbf{Synthetic data from the \autoref{sec:embeddings} on numerical embeddings}. For this dataset we tune the MLP-PLR model using the same protocol that is described in \autoref{sec-A:hyper-emp-details}.

\section{Technical details regarding synthetic dataset with $f(\cdot)$ and $g(\cdot)$ parameterized with MLPs}
\label{sec-A:synthetic-old}

We generate this dataset as follows. First, we sample 40000 20-dimentional standard normal vectors. Then, we employ two MLPs to parametrize functions $f(\cdot)$ and $g(\cdot)$. The MLP simulating function $f(\cdot)$ consists of three linear layers separated by ReLU activations. The MLP simulating function $g(\cdot)$ consists of two linear layers separated by a ReLU activation, with hidden dimension being equal to 10. We generate $y_i$ as $f(x_i) + g(x_i) \cdot \mathcal{N}(0, 1)$.

\section{Technical details regarding more effective embeddings learning and hyperparameter tuning}
\label{sec-A:scheme}

In this section we provide the hyperparameter tuning spaces for the more effective numerical features embeddings proposed in \autoref{sec:embeddings}. The tuning space is presented in \autoref{A:tab:mlp-lrlr-space}.

\begin{table}[h!]
\centering
\caption{The hyperparameter tuning space for the MLP and MLP-LRLR.}
{\renewcommand{\arraystretch}{1.1}
\begin{tabular}{lll}
    \toprule
    Parameter           & Distribution \\
    \midrule
    \# layers           & $\mathrm{UniformInt}[1,4]$ \\
    Width (hidden size) & $\mathrm{PowersOfTwo}[2^7,2^{11}]$ \\
    Dropout rate        & $\{0.0, \mathrm{Uniform}[0.0,0.75]\}$ \\
    Weight decay\textsuperscript{\scriptsize *}        & $\{0, \mathrm{LogUniform}[1e\text{-}6, 1e\text{-}3]\}$ \\
    Learning rate\textsuperscript{\scriptsize *}       & $\mathrm{LogUniform}[3e\text{-}5, 1e\text{-}3]$ \\
    \multicolumn{2}{l}{\scriptsize * -- We decouple optimizer parameters for the triplet pretraining and finetuning phases.} \\
    \midrule
    
    {\scriptsize \textbf{Embedding Parameters} (for LRLR and LRLR\textsubscript{triplet})}\vspace{4pt} \\ 
    d\_embedding      & $\{64, 128\}$ \\
    \# Tuning iterations & 100 \\
    \bottomrule
\end{tabular}}
\label{A:tab:mlp-lrlr-space}
\end{table}

The code for reproducing the results for the more effective embedding scheme with triplet-loss based pretraining is available in the supplementary materials. In \autoref{A:tab-all} we provide all unaggregated results on the regression datasets from the \citet{gorishniy2024tabm} benchmark no larger than 50K samples. We also exclude two problematic datasets reported by \citet{tschalzev2025unreflected}. When computing ranks for the aggregations above we use code provided with the supplementary materials that uses standard deviations to account for the statistical significance of the ranking.

\newcommand{\topalign}[1]{%
\vtop{\vskip 0pt #1}}

\begin{longtable}{p{0.5\textwidth}p{0.5\textwidth}}
\caption{Extended results for the main benchmark. Results are grouped by datasets.}
\label{A:tab-all}\\

\topalign{
\setlength\tabcolsep{2.5pt}
\renewcommand{\arraystretch}{0.9}
\begin{tabular}{lll}

\multicolumn{3}{c}{Ailerons \textdownarrow} \\
\toprule
Method & Single model & Ensemble \\
\midrule\\[-0.45cm]
\multicolumn{3}{c}{} \\[0.05cm]
{\footnotesize MLP } & {\footnotesize$0.0002 \pm 0.0000$} & {\footnotesize$0.0002 \pm 0.0000$}\\ 
{\footnotesize MLP-LRLR } & {\footnotesize$0.0002 \pm 0.0000$} & {\footnotesize$0.0002 \pm 0.0000$}\\ 
{\footnotesize MLP-LRLR\textsubscript{triplet} } & {\footnotesize$0.0002 \pm 0.0000$} & {\footnotesize$0.0002 \pm 0.0000$}\\ 
\bottomrule
\end{tabular}}

&

\topalign{
\setlength\tabcolsep{2.5pt}
\renewcommand{\arraystretch}{0.9}
\begin{tabular}{lll}

\multicolumn{3}{c}{MiamiHousing2016 \textdownarrow} \\
\toprule
Method & Single model & Ensemble \\
\midrule\\[-0.45cm]
\multicolumn{3}{c}{} \\[0.05cm]
{\footnotesize MLP } & {\footnotesize$0.1604 \pm 0.0029$} & {\footnotesize$0.1561 \pm 0.0026$}\\ 
{\footnotesize MLP-LRLR } & {\footnotesize$0.1507 \pm 0.0028$} & {\footnotesize$0.1483 \pm 0.0032$}\\ 
{\footnotesize MLP-LRLR\textsubscript{triplet} } & {\footnotesize$0.1471 \pm 0.0023$} & {\footnotesize$0.1445 \pm 0.0018$}\\ 
\bottomrule
\end{tabular}}

\\

\topalign{
\setlength\tabcolsep{2.5pt}
\renewcommand{\arraystretch}{0.9}
\begin{tabular}{lll}

\multicolumn{3}{c}{OnlineNewsPopularity \textdownarrow} \\
\toprule
Method & Single model & Ensemble \\
\midrule\\[-0.45cm]
\multicolumn{3}{c}{} \\[0.05cm]
{\footnotesize MLP } & {\footnotesize$0.8635 \pm 0.0007$} & {\footnotesize$0.8622 \pm 0.0003$}\\ 
{\footnotesize MLP-LRLR } & {\footnotesize$0.8595 \pm 0.0010$} & {\footnotesize$0.8568 \pm 0.0003$}\\ 
{\footnotesize MLP-LRLR\textsubscript{triplet} } & {\footnotesize$0.8581 \pm 0.0010$} & {\footnotesize$0.8557 \pm 0.0005$}\\ 
\bottomrule
\end{tabular}}

&

\topalign{
\setlength\tabcolsep{2.5pt}
\renewcommand{\arraystretch}{0.9}
\begin{tabular}{lll}

\multicolumn{3}{c}{analcatdata\_supreme \textdownarrow} \\
\toprule
Method & Single model & Ensemble \\
\midrule\\[-0.45cm]
\multicolumn{3}{c}{} \\[0.05cm]
{\footnotesize MLP } & {\footnotesize$0.0781 \pm 0.0097$} & {\footnotesize$0.0762 \pm 0.0090$}\\ 
{\footnotesize MLP-LRLR } & {\footnotesize$0.0775 \pm 0.0077$} & {\footnotesize$0.0762 \pm 0.0080$}\\ 
{\footnotesize MLP-LRLR\textsubscript{triplet} } & {\footnotesize$0.0774 \pm 0.0088$} & {\footnotesize$0.0766 \pm 0.0093$}\\ 
\bottomrule
\end{tabular}}

\\

\topalign{
\setlength\tabcolsep{2.5pt}
\renewcommand{\arraystretch}{0.9}
\begin{tabular}{lll}

\multicolumn{3}{c}{california \textdownarrow} \\
\toprule
Method & Single model & Ensemble \\
\midrule\\[-0.45cm]
\multicolumn{3}{c}{} \\[0.05cm]
{\footnotesize MLP } & {\footnotesize$0.4940 \pm 0.0048$} & {\footnotesize$0.4848 \pm 0.0015$}\\ 
{\footnotesize MLP-LRLR } & {\footnotesize$0.4628 \pm 0.0024$} & {\footnotesize$0.4536 \pm 0.0019$}\\ 
{\footnotesize MLP-LRLR\textsubscript{triplet} } & {\footnotesize$0.4582 \pm 0.0032$} & {\footnotesize$0.4494 \pm 0.0011$}\\ 
\bottomrule
\end{tabular}}

&

\topalign{
\setlength\tabcolsep{2.5pt}
\renewcommand{\arraystretch}{0.9}
\begin{tabular}{lll}

\multicolumn{3}{c}{cpu\_act \textdownarrow} \\
\toprule
Method & Single model & Ensemble \\
\midrule\\[-0.45cm]
\multicolumn{3}{c}{} \\[0.05cm]
{\footnotesize MLP } & {\footnotesize$2.7117 \pm 0.1624$} & {\footnotesize$2.5668 \pm 0.1102$}\\ 
{\footnotesize MLP-LRLR } & {\footnotesize$2.2692 \pm 0.0676$} & {\footnotesize$2.1990 \pm 0.0698$}\\ 
{\footnotesize MLP-LRLR\textsubscript{triplet} } & {\footnotesize$2.2782 \pm 0.1183$} & {\footnotesize$2.2019 \pm 0.0979$}\\ 
\bottomrule
\end{tabular}}

\\

\topalign{
\setlength\tabcolsep{2.5pt}
\renewcommand{\arraystretch}{0.9}
\begin{tabular}{lll}

\multicolumn{3}{c}{diamond \textdownarrow} \\
\toprule
Method & Single model & Ensemble \\
\midrule\\[-0.45cm]
\multicolumn{3}{c}{} \\[0.05cm]
{\footnotesize MLP } & {\footnotesize$0.1396 \pm 0.0012$} & {\footnotesize$0.1370 \pm 0.0006$}\\ 
{\footnotesize MLP-LRLR } & {\footnotesize$0.1342 \pm 0.0013$} & {\footnotesize$0.1324 \pm 0.0008$}\\ 
{\footnotesize MLP-LRLR\textsubscript{triplet} } & {\footnotesize$0.1328 \pm 0.0010$} & {\footnotesize$0.1321 \pm 0.0004$}\\ 
\bottomrule
\end{tabular}}

&

\topalign{
\setlength\tabcolsep{2.5pt}
\renewcommand{\arraystretch}{0.9}
\begin{tabular}{lll}

\multicolumn{3}{c}{elevators \textdownarrow} \\
\toprule
Method & Single model & Ensemble \\
\midrule\\[-0.45cm]
\multicolumn{3}{c}{} \\[0.05cm]
{\footnotesize MLP } & {\footnotesize$0.0020 \pm 0.0000$} & {\footnotesize$0.0019 \pm 0.0000$}\\ 
{\footnotesize MLP-LRLR } & {\footnotesize$0.0018 \pm 0.0000$} & {\footnotesize$0.0018 \pm 0.0000$}\\ 
{\footnotesize MLP-LRLR\textsubscript{triplet} } & {\footnotesize$0.0018 \pm 0.0000$} & {\footnotesize$0.0018 \pm 0.0000$}\\ 
\bottomrule
\end{tabular}}

\\

\topalign{
\setlength\tabcolsep{2.5pt}
\renewcommand{\arraystretch}{0.9}
\begin{tabular}{lll}

\multicolumn{3}{c}{fifa \textdownarrow} \\
\toprule
Method & Single model & Ensemble \\
\midrule\\[-0.45cm]
\multicolumn{3}{c}{} \\[0.05cm]
{\footnotesize MLP } & {\footnotesize$0.8025 \pm 0.0135$} & {\footnotesize$0.8005 \pm 0.0149$}\\ 
{\footnotesize MLP-LRLR } & {\footnotesize$0.7880 \pm 0.0114$} & {\footnotesize$0.7849 \pm 0.0122$}\\ 
{\footnotesize MLP-LRLR\textsubscript{triplet} } & {\footnotesize$0.7863 \pm 0.0112$} & {\footnotesize$0.7836 \pm 0.0124$}\\ 
\bottomrule
\end{tabular}}

&

\topalign{
\setlength\tabcolsep{2.5pt}
\renewcommand{\arraystretch}{0.9}
\begin{tabular}{lll}

\multicolumn{3}{c}{house \textdownarrow} \\
\toprule
Method & Single model & Ensemble \\
\midrule\\[-0.45cm]
\multicolumn{3}{c}{} \\[0.05cm]
{\footnotesize MLP } & {\footnotesize$3.1105 \pm 0.0439$} & {\footnotesize$3.0385 \pm 0.0390$}\\ 
{\footnotesize MLP-LRLR } & {\footnotesize$3.1181 \pm 0.0568$} & {\footnotesize$3.0434 \pm 0.0354$}\\ 
{\footnotesize MLP-LRLR\textsubscript{triplet} } & {\footnotesize$3.0705 \pm 0.0156$} & {\footnotesize$3.0470 \pm 0.0078$}\\ 
\bottomrule
\end{tabular}}

\\

\topalign{
\setlength\tabcolsep{2.5pt}
\renewcommand{\arraystretch}{0.9}
\begin{tabular}{lll}

\multicolumn{3}{c}{house\_sales \textdownarrow} \\
\toprule
Method & Single model & Ensemble \\
\midrule\\[-0.45cm]
\multicolumn{3}{c}{} \\[0.05cm]
{\footnotesize MLP } & {\footnotesize$0.1812 \pm 0.0009$} & {\footnotesize$0.1781 \pm 0.0004$}\\ 
{\footnotesize MLP-LRLR } & {\footnotesize$0.1677 \pm 0.0005$} & {\footnotesize$0.1660 \pm 0.0002$}\\ 
{\footnotesize MLP-LRLR\textsubscript{triplet} } & {\footnotesize$0.1696 \pm 0.0006$} & {\footnotesize$0.1677 \pm 0.0001$}\\ 
\bottomrule
\end{tabular}}

&

\topalign{
\setlength\tabcolsep{2.5pt}
\renewcommand{\arraystretch}{0.9}
\begin{tabular}{lll}

\multicolumn{3}{c}{isolet \textdownarrow} \\
\toprule
Method & Single model & Ensemble \\
\midrule\\[-0.45cm]
\multicolumn{3}{c}{} \\[0.05cm]
{\footnotesize MLP } & {\footnotesize$2.2740 \pm 0.3170$} & {\footnotesize$2.0363 \pm 0.1421$}\\ 
{\footnotesize MLP-LRLR } & {\footnotesize$2.3179 \pm 0.1404$} & {\footnotesize$2.1286 \pm 0.0940$}\\ 
{\footnotesize MLP-LRLR\textsubscript{triplet} } & {\footnotesize$2.3119 \pm 0.1506$} & {\footnotesize$2.1873 \pm 0.1349$}\\ 
\bottomrule
\end{tabular}}

\\

\topalign{
\setlength\tabcolsep{2.5pt}
\renewcommand{\arraystretch}{0.9}
\begin{tabular}{lll}

\multicolumn{3}{c}{particulate-matter-ukair-2017 \textdownarrow} \\
\toprule
Method & Single model & Ensemble \\
\midrule\\[-0.45cm]
\multicolumn{3}{c}{} \\[0.05cm]
{\footnotesize MLP } & {\footnotesize$0.3779 \pm 0.0006$} & {\footnotesize$0.3754 \pm 0.0002$}\\ 
{\footnotesize MLP-LRLR } & {\footnotesize$0.3661 \pm 0.0009$} & {\footnotesize$0.3634 \pm 0.0002$}\\ 
{\footnotesize MLP-LRLR\textsubscript{triplet} } & {\footnotesize$0.3652 \pm 0.0006$} & {\footnotesize$0.3629 \pm 0.0002$}\\ 
\bottomrule
\end{tabular}}

&

\topalign{
\setlength\tabcolsep{2.5pt}
\renewcommand{\arraystretch}{0.9}
\begin{tabular}{lll}

\multicolumn{3}{c}{pol \textdownarrow} \\
\toprule
Method & Single model & Ensemble \\
\midrule\\[-0.45cm]
\multicolumn{3}{c}{} \\[0.05cm]
{\footnotesize MLP } & {\footnotesize$5.6147 \pm 0.6212$} & {\footnotesize$5.1092 \pm 0.6127$}\\ 
{\footnotesize MLP-LRLR } & {\footnotesize$2.6721 \pm 0.1627$} & {\footnotesize$2.4205 \pm 0.1062$}\\ 
{\footnotesize MLP-LRLR\textsubscript{triplet} } & {\footnotesize$2.5144 \pm 0.1259$} & {\footnotesize$2.3130 \pm 0.0560$}\\ 
\bottomrule
\end{tabular}}

\\

\topalign{
\setlength\tabcolsep{2.5pt}
\renewcommand{\arraystretch}{0.9}
\begin{tabular}{lll}

\multicolumn{3}{c}{sberbank-housing \textdownarrow} \\
\toprule
Method & Single model & Ensemble \\
\midrule\\[-0.45cm]
\multicolumn{3}{c}{} \\[0.05cm]
{\footnotesize MLP } & {\footnotesize$0.2508 \pm 0.0046$} & {\footnotesize$0.2447 \pm 0.0019$}\\ 
{\footnotesize MLP-LRLR } & {\footnotesize$0.2427 \pm 0.0054$} & {\footnotesize$0.2383 \pm 0.0014$}\\ 
{\footnotesize MLP-LRLR\textsubscript{triplet} } & {\footnotesize$0.2405 \pm 0.0084$} & {\footnotesize$0.2323 \pm 0.0013$}\\ 
\bottomrule
\end{tabular}}

&

\topalign{
\setlength\tabcolsep{2.5pt}
\renewcommand{\arraystretch}{0.9}
\begin{tabular}{lll}

\multicolumn{3}{c}{superconduct \textdownarrow} \\
\toprule
Method & Single model & Ensemble \\
\midrule\\[-0.45cm]
\multicolumn{3}{c}{} \\[0.05cm]
{\footnotesize MLP } & {\footnotesize$10.7537 \pm 0.0778$} & {\footnotesize$10.3687 \pm 0.0162$}\\ 
{\footnotesize MLP-LRLR } & {\footnotesize$10.7919 \pm 0.1539$} & {\footnotesize$10.3938 \pm 0.0130$}\\ 
{\footnotesize MLP-LRLR\textsubscript{triplet} } & {\footnotesize$10.6695 \pm 0.0817$} & {\footnotesize$10.3082 \pm 0.0451$}\\ 
\bottomrule
\end{tabular}}

\\

\topalign{
\setlength\tabcolsep{2.5pt}
\renewcommand{\arraystretch}{0.9}
\begin{tabular}{lll}

\multicolumn{3}{c}{wine\_quality \textdownarrow} \\
\toprule
Method & Single model & Ensemble \\
\midrule\\[-0.45cm]
\multicolumn{3}{c}{} \\[0.05cm]
{\footnotesize MLP } & {\footnotesize$0.6682 \pm 0.0139$} & {\footnotesize$0.6573 \pm 0.0153$}\\ 
{\footnotesize MLP-LRLR } & {\footnotesize$0.6717 \pm 0.0155$} & {\footnotesize$0.6510 \pm 0.0170$}\\ 
{\footnotesize MLP-LRLR\textsubscript{triplet} } & {\footnotesize$0.6770 \pm 0.0212$} & {\footnotesize$0.6608 \pm 0.0229$}\\ 
\bottomrule
\end{tabular}}

&

\topalign{
\setlength\tabcolsep{2.5pt}
\renewcommand{\arraystretch}{0.9}
\begin{tabular}{lll}

\multicolumn{3}{c}{year \textdownarrow} \\
\toprule
Method & Single model & Ensemble \\
\midrule\\[-0.45cm]
\multicolumn{3}{c}{} \\[0.05cm]
{\footnotesize MLP } & {\footnotesize$8.9732 \pm 0.0237$} & {\footnotesize$8.8923 \pm 0.0058$}\\ 
{\footnotesize MLP-LRLR } & {\footnotesize$8.9511 \pm 0.0190$} & {\footnotesize$8.9168 \pm 0.0059$}\\ 
{\footnotesize MLP-LRLR\textsubscript{triplet} } & {\footnotesize$8.9398 \pm 0.0095$} & {\footnotesize$8.9246 \pm 0.0042$}\\ 
\bottomrule
\end{tabular}}

\\
\end{longtable}

\section{Consistency of the provided results when using different models used for data uncertainty estimation}
\label{sec-A:consistency}

In this section, we demonstrate that results we obtained in \autoref{sec:embeddings}, \autoref{sec:mnca} and \autoref{sec:tabm} stay the same when using different models for uncertainty estimation.

\begin{figure*}[h!]
    \centering
    \includegraphics[width=\linewidth]{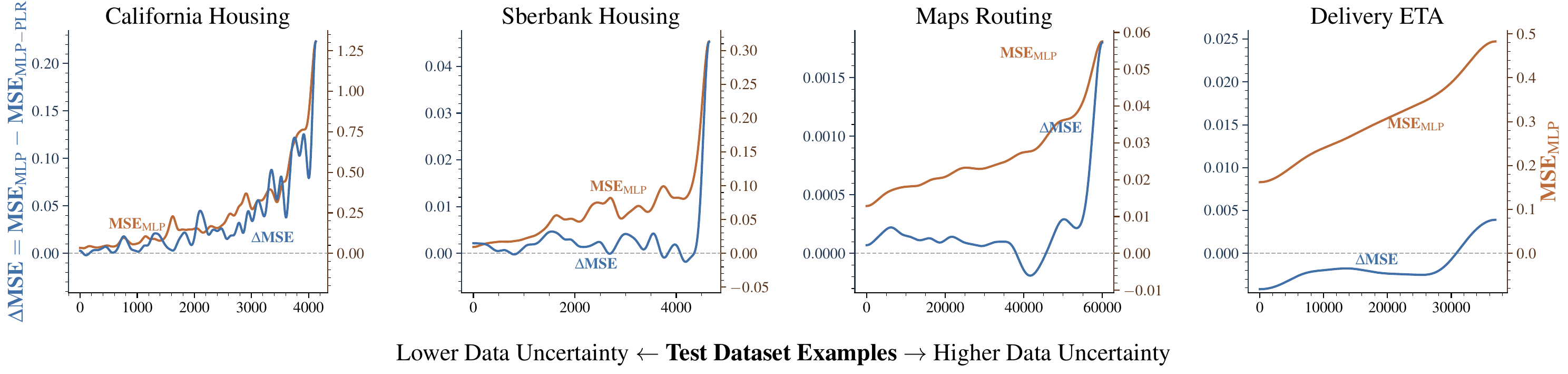}
    \caption{Version of \autoref{fig:plr-compare}, but using MLP as a basic model for estimating data uncertainty. As can be seen, using different model for estimation leads to the same conclusion.}
    \label{fig:plr_compare_by_mlp}
\end{figure*}

\begin{figure*}[h!]
    \centering
    \includegraphics[width=\linewidth]{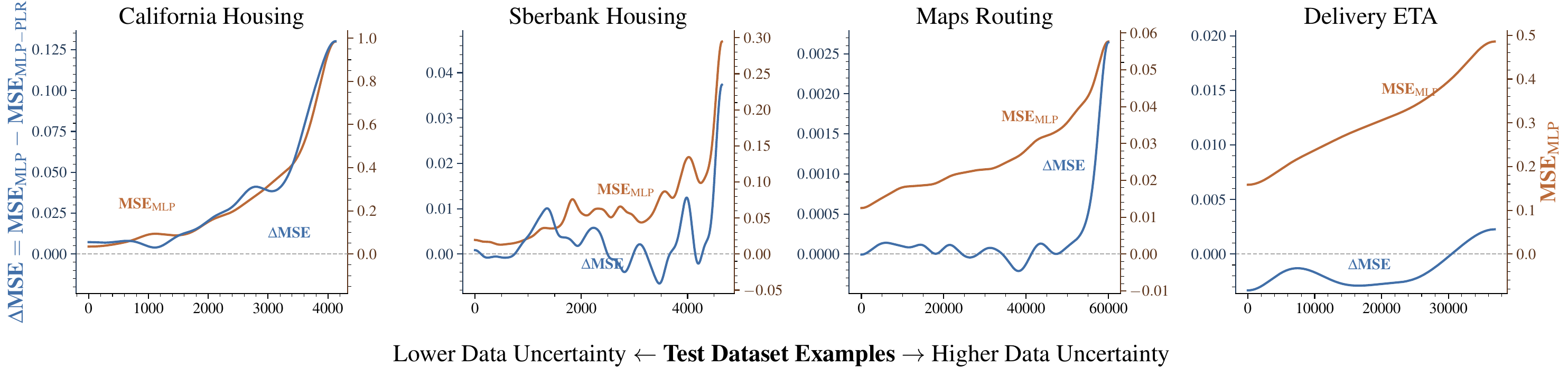}
    \caption{Version of \autoref{fig:plr-compare}, but using MLP-PLR as a basic model for estimating data uncertainty. As can be seen, using different model for estimation leads to the same conclusion.}
    \label{fig:plr_compare_by_mlp_plr}
\end{figure*}

\begin{figure*}[h!]
    \centering
    \includegraphics[width=\linewidth]{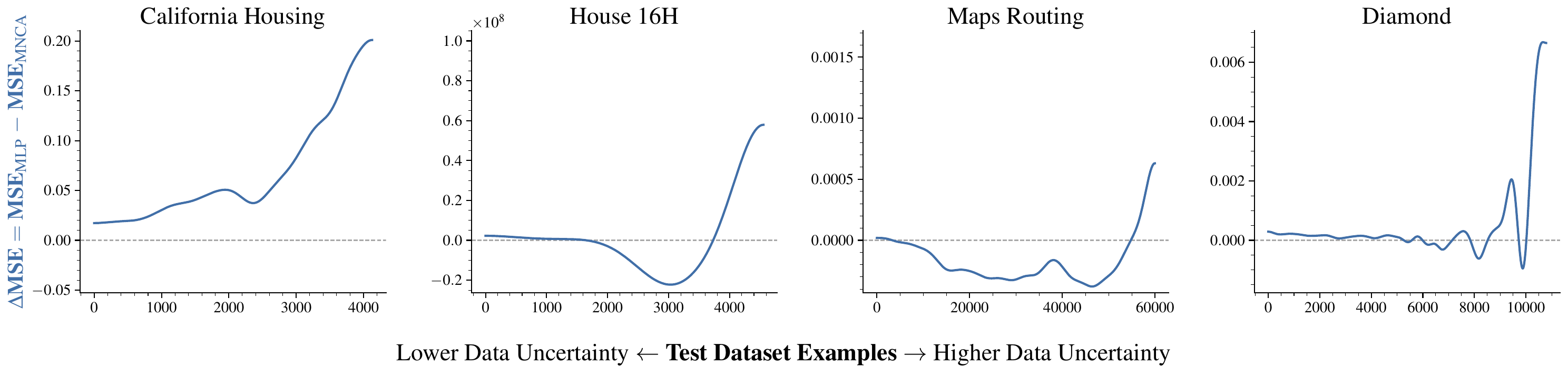}
    \caption{Version of \autoref{fig:mnca-compare}, but using MLP as a basic model for estimating data uncertainty. As can be seen, using different model for estimation leads to the same conclusion.}
    \label{fig:mnca_compare_by_mlp}
\end{figure*}

\begin{figure*}[h!]
    \centering
    \includegraphics[width=\linewidth]{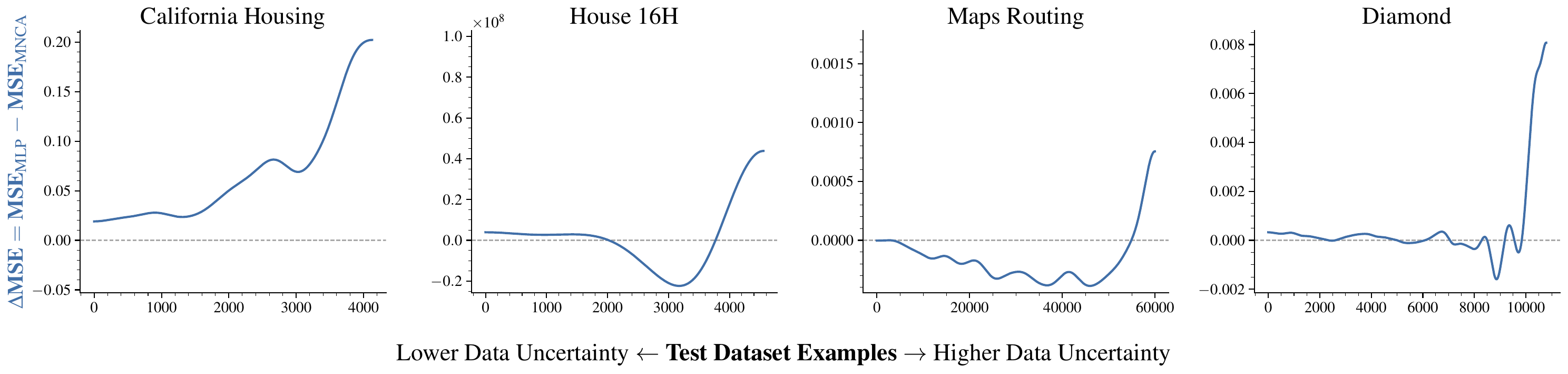}
    \caption{Version of \autoref{fig:mnca-compare}, but using MLP-PLR as a basic model for estimating data uncertainty. As can be seen, using different model for estimation leads to the same conclusion.}
    \label{fig:mnca_compare_by_mlp_plr}
\end{figure*}

\begin{figure*}[h!]
    \centering
    \includegraphics[width=\linewidth]{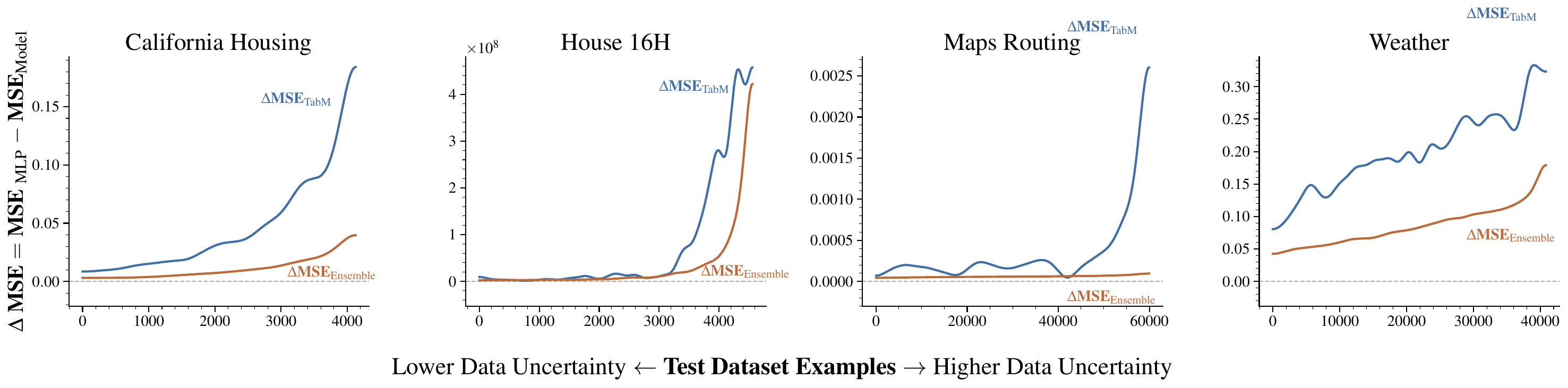}
    \caption{Version of \autoref{fig:tabm_compare}, but using MLP as a basic model for estimating data uncertainty. As can be seen, using different model for estimation leads to the same conclusion.}
    \label{fig:tabm_compare_by_mlp}
\end{figure*}

\begin{figure*}[h!]
    \centering
    \includegraphics[width=\linewidth]{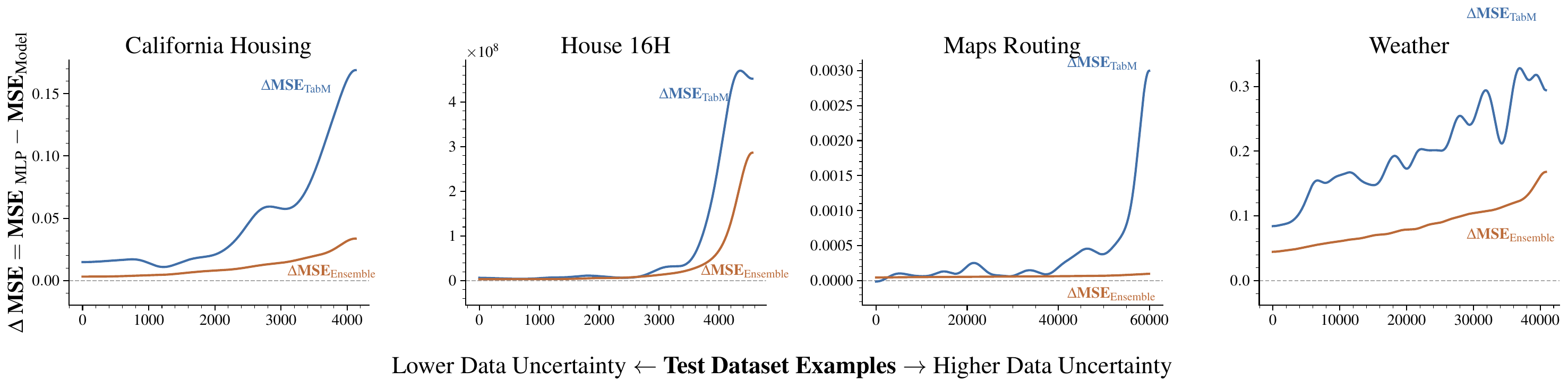}
    \caption{Version of \autoref{fig:tabm_compare}, but using MLP-PLR as a basic model for estimating data uncertainty. As can be seen, using different model for estimation leads to the same conclusion.}
    \label{fig:tabm_compare_by_mlp_plr}
\end{figure*}

As can be seen on \autoref{fig:plr_compare_by_mlp}, \autoref{fig:plr_compare_by_mlp_plr}, \autoref{fig:mnca_compare_by_mlp}, \autoref{fig:mnca_compare_by_mlp_plr}, \autoref{fig:tabm_compare_by_mlp}, \autoref{fig:tabm_compare_by_mlp_plr}, our analysis is not dependent on which specific model we use to estimate data uncertainty.

\end{document}